\documentclass[10pt,twocolumn,letterpaper]{article}

\usepackage{cvpr}      

\usepackage{graphicx}
\usepackage{amsmath}
\usepackage{amssymb}
\usepackage{booktabs}
\usepackage{blindtext}
\usepackage{pifont}
\usepackage{multirow, makecell}
\usepackage[pagebackref,breaklinks,colorlinks]{hyperref}

\usepackage[capitalize]{cleveref}

\usepackage{xcolor}
\usepackage{tabularx}
\usepackage[utf8]{inputenc}
\DeclareUnicodeCharacter{2014}{\cross}

\crefname{section}{Sec.}{Secs.}
\Crefname{section}{Section}{Sections}
\Crefname{table}{Table}{Tables}
\crefname{table}{Tab.}{Tabs.}

\newcommand{\paragraphneurips}[1]{

\paragraph{#1}
}

\newcommand\blfootnote[1]{%
  \begingroup
  \renewcommand\thefootnote{}\footnote{\noindent#1}%
  \addtocounter{footnote}{-1}%
  \endgroup
}

\usepackage{tikz}

\usepackage[precision=2, unit=mm]{lengthconvert}
\usetikzlibrary{patterns}

\def\confName{CVPR}
\def\confYear{2024}

\newcommand{\model}{\mbox{\textsc{LEdits++}}}

\crefformat{footnote}{#2\footnotemark[#1]#3}

\begin{document}

\title{\textsc{LEdits++}: Limitless Image Editing using Text-to-Image Models}

\author{ 
Manuel Brack$^{1,2*}$\textsuperscript{\textdagger} \and Felix Friedrich$^{1,3*}$ \and Katharina Kornmeier$^{1*}$ \and Linoy Tsaban$^{4}$ \and Patrick Schramowski$^{1,2,3}$ \and Kristian Kersting$^{1,2,3}$ \and Apolinário Passos$^{4}$ \and \vspace{-4mm}\\
\text{$\phantom{0}^{1}$TU Darmstadt,$\phantom{0}^{2}$DFKI,$\phantom{0}^{3}$hessian.AI,$\phantom{0}^{4}$Huggingface} \\
{\tt\small \{brack,friedrich\}@cs.tu-darmstadt.de}
}

\twocolumn[{%
\renewcommand\twocolumn[1][]{#1}%
\maketitle

\begin{center}

    \centering
    \includegraphics[width=.85\linewidth]{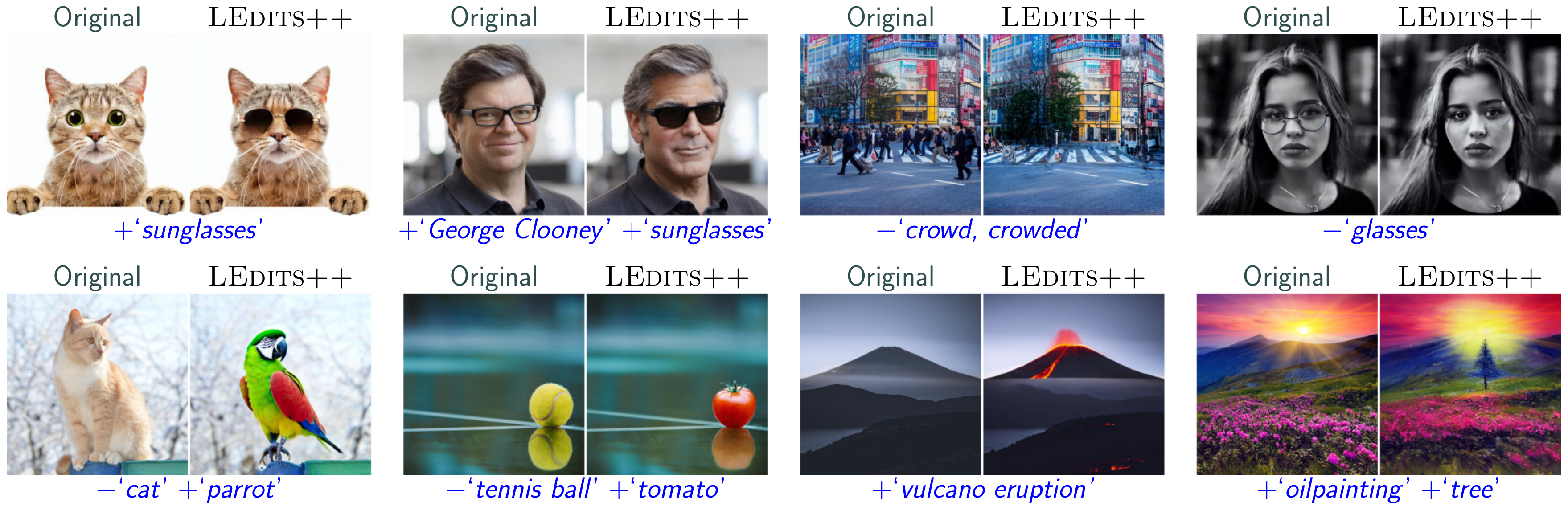}
    \vskip -.5em
    \captionof{figure}{\model~facilitates versatile image-to-image editing. Several complex cases are available now.}
    \label{fig:basic_examples}
    \vskip 1.5em
\end{center}
}]


\begin{abstract}

\blfootnote{*Equal contribution \qquad \textdagger Partially as research intern at Adobe}
\blfootnote{Proceedings of the 2024 IEEE/CVF Conference on Computer Vision and Pattern Recognition (CVPR)}
\noindent Text-to-image diffusion models have recently received increasing interest for their astonishing ability to produce high-fidelity images from solely text inputs. Subsequent research efforts aim to exploit and apply their capabilities to real image editing. However, existing image-to-image methods are often inefficient, imprecise, and of limited versatility. They either require time-consuming fine-tuning, deviate unnecessarily strongly from the input image, and/or lack support for multiple, simultaneous edits. To address these issues, we introduce \model, an efficient yet versatile and precise textual image manipulation technique. 
\model's novel inversion approach requires no tuning nor optimization and produces high-fidelity results with a few diffusion steps. Second, our methodology supports multiple simultaneous edits and is architecture-agnostic. Third, we use a novel implicit masking technique that limits changes to relevant image regions. We propose the novel \textit{TEdBench++} benchmark as part of our exhaustive evaluation. Our results demonstrate the capabilities of \model~and its improvements over previous methods.
\end{abstract}

\section{Introduction}\label{sec:intro}\noindent
Text-to-image diffusion models (DM) have garnered recognition for their ability to generate high-quality images from textual descriptions. A growing body of research has recently been dedicated to utilizing these models for manipulating real images. 

However, several barriers prevent many real-world applications of diffusion-based image editing. Current methods often entail computationally expensive model tuning or other optimization, presenting practical challenges \cite{kawar2023imagic, valevski2022UniTune, mokady2023nulltext, brooks2022instructpix2pix, parmar2023zeroshot}. Additionally, existing techniques have the proclivity to induce profound changes to the original image \cite{meng2022sdedit, hubermanspiegelglas2023edit}, often resulting in completely different images. Lastly, all these approaches are inherently constrained when editing multiple (arbitrary) concepts simultaneously.
We tackle these problems by introducing \model\footnote{\textsc{\model} stands for \textit{\textbf{L}imitless \textbf{Edits}} with sde-dpm-solver\textbf{++}.}, a diffusion-based image editing technique addressing these limitations.

\model\footnote{\label{note1}{\scriptsize\url{https://huggingface.co/spaces/leditsplusplus/project}}}~offers a streamlined approach for textual image editing, eliminating the need for extensive parameter tuning. To this end, we derive image inversion for a more efficient diffusion sampling algorithm to a) drastically reduce computational resources and b) guarantee perfect image reconstruction. Thus, we overcome computational obstacles and avoid changes in the edited image in the first place. Furthermore, we use a novel implicit masking approach to semantically ground each edit instruction to its relevant image region. This further optimizes changes to the image by retaining the overall image composition and object identity.
Additionally, \model~is the only method to date to facilitate easy and versatile image editing by supporting multiple simultaneous instructions without causing undue interference. 
Finally, its lightweight architecture-agnostic nature ensures compatibility with both latent and pixel-based diffusion models, providing high accessibility.

In this work, we establish the methodical benefits of \model~and demonstrate that this intuitive,
lightweight approach offers sophisticated semantic control for image editing. Specifically,
we contribute by (i) devising a formal definition of \model~while (ii) deriving perfect inversion for a more efficient diffusion sampling method, (iii) qualitatively and empirically demonstrating its efficiency, versatility, and precision, (iv) providing an exhaustive empirical comparison to concurrent works with automatic and human user metrics,
and thereby (v) introducing \textbf{T}extual \textbf{Ed}iting \textbf{B}enchmark++ (TEdBench++), a more holistic and coherent testbed for evaluating textual image manipulation.


    
\section{Background}\label{sec:rel_work} \noindent
%
Recently, large-scale, text-guided DMs have enabled versatile applications in image generation
\cite{saharia2022photorealistic, ramesh2022hierarchical, balaji2022ediffi}. Especially latent diffusion models \cite{rombach2022High, pernias2023wuerstchen} have gained  attention for their computational efficiency.
Below, we discuss related work for efficient, versatile image manipulation with DMs.

\paragraphneurips{Diffusion Sampling.}
Generating outputs with DMs requires multiple iterative denoising steps that constitute the main bottleneck at inference. Commonly used sampling methods such as DDPM \cite{ho2020denoising} or DDIM \cite{song2021denoising} require tens or hundreds of steps to produce high-quality samples. Consequently, numerous works have been dedicated to speeding up the sampling process without loss in quality. Distillation efforts progressively reduce the number of required steps through further training \cite{meng2023ondistillation, luo2023latent}. 
Other works focus on improving the sampling itself, e.g.~using high-order ODE-solvers \cite{lu2022dpmsolver, lu2023dpmsolver, zhao2023unipc}. Such solvers can be readily combined with pre-trained DMs at inference to lower the number of denoising steps. With \model, we derive perfect image inversion with the DPM-Solver++, allowing image editing in as few as 20 total steps.

\paragraphneurips{Semantic Control during Diffusion.}
While text-to-image DMs generate new, astonishing images, fine-grained control over the generative process remains challenging. Minor changes to the text prompt lead to entirely different outputs. Wu et al.~\cite{wu2022uncovering} studied concept disentanglement using linear combinations of text embeddings to gain semantic control. 
Methods like Prompt-to-Prompt \cite{hertz2022prompt} and other works \cite{chefer2023attend,parmar2023zeroshot}
utilize 
the DM's attention layers to attribute pixels to tokens from the text prompt. Dedicated operations on the attention maps enable more control over the generated images.
Other works have focused on the noise estimates of DMs \cite{brack2023Sega,liu2022Compositional} 
providing semantic control over the generation process.
With \model, we now enable fine-grained semantic control for manipulating real images, going beyond purely generative applications.



\paragraphneurips{Real Image Editing.}
Since DMs' rise in popularity for text-to-image generation, they have also been explored for (real) image-to-image editing.
As a first, simple approach, SDEdit added noise to the image for an intermediate step in the diffusion process \cite{meng2022sdedit}. While lightweight, the resulting image diverges substantially from the input as it is (partially) regenerated. Inpainting allows to keep the change small by having a user provide additional masks to restrict changes to certain image regions \cite{avrahmi2022blended, nichol2022glide}.
Yet, user masks are costly or often simply unavailable. 
Other works have thus explored semantically grounded approaches using cross-attention instead to better control image manipulation \cite{cao2023masactrl, parmar2023zeroshot, mokady2023nulltext}. 
In contrast, \model~leverages both attention- and noise-based masking to obtain fine-grained masks, enabling strong semantic control over real images.

Another important aspect of image manipulation methods is the required tuning and overall runtime. InstructPix2Pix continues training a DM at scale to enable image editing capabilities \cite{brooks2022instructpix2pix}. 
Finetuning instead on each individual input to constrain the generation on the real image has shown helpful \cite{kawar2023imagic, valevski2022UniTune} but not computationally efficient. 
Consequently, recent works have largely relied on inverting the deterministic DDIM sampling process \cite{song2021denoising} to save computational resources. DDIM inversion identifies an initial noise vector that results in the input image when denoised again. However, faithful reconstructions are only obtained in the limit of small steps, thus requiring large numbers of inversion steps. Moreover, small errors will still incur at each timestep, often accumulating into meaningful deviations from the input, requiring costly error correction through optimization \cite{parmar2023zeroshot, mokady2023nulltext}. 
Recently, Huberman-Spiegelglas \mbox{\it et al.}~proposed an inversion technique \cite{hubermanspiegelglas2023edit} for the DDPM sampler \cite{ho2020denoising} to address the limitations of DDIM inversion. 
\model~provides the same guarantees of perfect inversion with even further reduced runtime alongside an edit-friendly latent space, enabling more versatility.



\section{Image Editing with Text-to-Image Models}\noindent
Before devising the methodology of \model, let us first motivate the desired features and use cases. Specifically, we aim for efficiency, versatility, and precision. The goal is to provide a method that enables a fast exploratory workflow for image editing in which a user can iteratively interact with the model and explore various edits. Consequently, \model~produces outputs quickly with no tuning or optimization to not disrupt the creative process. Further, arbitrary editing instructions and combinations thereof are supported to facilitate a wide range of image manipulations (e.g.,~complex multi-editing). 
Lastly, we provide precise and sophisticated semantic control over the image editing. Each of the (potentially multiple) edit instructions can be steered individually, and changes are automatically restricted to relevant image regions. Importantly, with \model~we prioritize compositional robustness. 



\subsection{Guided Diffusion Models}\noindent
Let us first define some general background for diffusion models.
DMs iteratively denoise a Gaussian distributed variable to produce samples of a learned data distribution. Let's consider a diffusion process that gradually turns an image $x_0$ into Gaussian noise. 
\begin{equation}
    x_t = \sqrt{1-\beta_t}x_{t-1} + \sqrt{\beta_t} n_t, \qquad t= 1,...,T
\end{equation}
where $n_t$ are iid normal distributed vectors and $\beta_t$ a variance schedule. The diffusion process is equivalently expressed as 
\begin{equation}
    x_t = \sqrt{\bar{\alpha}_t} x_0 + \sqrt{1- \bar{\alpha}_t}\epsilon_t
    \label{eq:xt}
\end{equation}
where $\alpha_t = 1 - \beta_t$, $\bar{\alpha}_t= \Pi^t_{s=1}\alpha_s$ and $\epsilon_t\sim\mathcal{N}(0,\mathbf{I})$. 
Importantly, all $\epsilon_t$ are \textit{not} statistically independent. Instead, consecutive pairs $\epsilon_t, \epsilon_{t-1}$ are strongly dependent, which will be relevant later.
To generate an (new) image $\hat{x}_0$ the reverse diffusion process starts from random noise $x_T \sim \mathcal{N}(0,\mathbf{I})$ which can be iteratively denoised as
\begin{equation}
    x_{t-1} = \hat{\mu}_t(x_t) + \sigma_t z_t, \qquad t= T,...,1
    \label{eq:denoise}
\end{equation}
Here $z_t$ are iid standard normal vectors, and common variance schedulers $\sigma_t$ can be expressed in the general form
$$
    \sigma_t= \eta \beta_t \frac{1-\bar{\alpha}_{t-1}}{1-\bar{\alpha}_{t}}
$$
where $\eta \in [0,1]$. In this formulation, $\eta = 0$ corresponds to the deterministic DDIM \cite{song2021denoising} and $\eta = 1$ to the DDPM scheme \cite{ho2020denoising}. 
Lastly, in theses cases, we have $\hat{\mu}_t(x_t)=$ 
$$
\sqrt{\bar{\alpha}_{t-1}} \frac{x_t -\sqrt{1-\bar{\alpha}_{t} }\hat{\epsilon}_\theta(x_t)}{\sqrt{\bar{\alpha}_{t}}}
    + \sqrt{1-\bar{\alpha}_{t-1} - \sigma^2_t}\hat{\epsilon}_\theta(x_t)
$$
Here $\hat{\epsilon}_\theta(x_t)$ is an estimate of $\epsilon_t$ produced by our neural network DM with learned parameters $\theta$, commonly implemented as a U-Net \cite{ronneberger2015unet}.
For text-to-image generation, the model is conditioned on a text prompt $p$ to produce images faithful to that prompt. 
The DM is trained to produce the noise estimate $\hat{\epsilon}_\theta(x_t)$ needed for iteratively sampling $\hat{x}_0$ (Eq.~\ref{eq:denoise}). For text-conditioned DMs, $\hat{\epsilon}_\theta$ is calculated using specific guidance techniques. 
%
%
%
%

Most DMs rely on classifier-free guidance \cite{ho2022classifier}, a conditioning method using a purely generative diffusion model, eliminating the need for an additional classifier. During training, the text conditioning $c_p$ is randomly dropped with a fixed probability, resulting in a joint model for unconditional and conditional objectives. 
During inference, the score estimates for the $\epsilon$-prediction are adjusted so that: 
\begin{equation}\label{eq:classifier_free}
    \hat{\epsilon}_\theta({x}_t, c_p) := \hat{\epsilon}_\theta({x}_t) + s_g (\hat{\epsilon}_\theta({x}_t, c_p) - \hat{\epsilon}_\theta({x}_t))
\end{equation}
with guidance scale $s_g$ and $\hat{\epsilon}_\theta$ defining the noise estimate with parameters $\theta$. Intuitively, the unconditioned $\epsilon$-prediction is pushed in the direction of the conditioned one, with $s_g$ determining the extent of the adjustment.

\subsection{\textbf{\textsc{LEdits}++}}
\label{subsec:ledits}\noindent
With the fundamentals established, the methodology of \model~can now be broken down into three components: (1) efficient image inversion, (2) versatile textual editing, and (3) semantic grounding of image changes. 

\paragraphneurips{Component 1: Perfect Inversion.}
Utilizing text-to-image models for editing real images requires conditioning the generation on the input image. Recent works have largely relied on inverting the sampling process to identify $x_T$ that will be denoised to the input image $x_0$ \cite{mokady2023nulltext, parmar2023zeroshot}. Inverting the DDPM scheduler is generally preferred over DDIM inversion since the former can be achieved in fewer timesteps and with no reconstruction error \cite{hubermanspiegelglas2023edit}.

However, there exist more efficient schemes than DDPM for sampling DMs that greatly reduce the required number of steps and consequently DM evaluations. We here propose a more efficient inversion method by deriving the desired inversion properties for such a scheme.
As demonstrated by Song \mbox{\it et al.}\cite{Song2021score}, DDPM can be viewed as a first-order stochastic differential equation (SDE) solver when formulating the reverse diffusion process as an SDE. This SDE can be solved more efficiently---in fewer steps---using a higher-order differential equation solver, here \textit{dpm-solver++} \cite{lu2023dpmsolver}. The reverse diffusion process from Eq.~\ref{eq:denoise} for the second-order sde-dpm-solver++ can be written as  
\begin{equation}
    x_{t-1} = \hat{\mu}_t(x_t, x_{t+1}) + \sigma_t z_t, \qquad t= T,...,1
    \label{eq:denoise_plus}
\end{equation}
where now
$$
    \sigma_t =  \sqrt{1-\bar{\alpha}_{t-1}} \sqrt{1-e^{-2h_{t-1}}} 
$$
and higher-order $\hat{\mu}_t$ depends now on $x$ from two timesteps, $x_t$ \textit{and} $x_{t+1}$. Such that $\hat{\mu}_t(x_t, x_{t+1}) =$
\begin{align}
\nonumber    
    &\frac{\sqrt{1-\bar{\alpha}_{t-1} }}{\sqrt{1-\bar{\alpha}_{t}}} e^{-h_{t-1}} x_t + \sqrt{\bar{\alpha}_{t-1}} \big(1-e^{-2h_{t-1}}\big) \hat{\epsilon}_\theta({x}_t) \\ \nonumber
    &+ 0.5 \sqrt{\bar{\alpha}_{t-1}}\big(1-e^{-2h_{t-1}}\big) \frac{-h_t}{h_{t-1}} \big(\hat{\epsilon}_\theta({x}_{t+1}) -  \hat{\epsilon}_\theta({x}_t)\big)    
\end{align}
with 
$$h_t = \frac{\ln(\sqrt{\bar{\alpha}_{t}})}{\ln(\sqrt{1-\bar{\alpha}_{t}})} - \frac{\ln(\sqrt{\bar{\alpha}_{t+1}}) }{\ln(\sqrt{1-\bar{\alpha}_{t+1}})}  $$
For the detailed derivation of the solver and proof of faster convergence, we refer the reader to the relevant literature \cite{lu2023dpmsolver, lu2022dpmsolver}. 
Based on the above, we now devise our inversion process. Given an input image $x_0$ we construct an auxiliary reconstruction sequence of noise images $x_1,...,x_T$ as
\begin{equation}
    x_t = \sqrt{\bar{\alpha}_t} x_0 + \sqrt{1- \bar{\alpha}_t}\tilde{\epsilon}_t
\end{equation}
where $\tilde{\epsilon}_t\sim\mathcal{N}(0,\mathbf{I})$. Contrary to Eq.~\ref{eq:xt}, the $\tilde{\epsilon}_t$ are now statistically \textit{independent}, which is a desirable property for image editing \cite{hubermanspiegelglas2023edit}. Lastly, the respective $z_t$ for the inversion can be derived from Eq.~\ref{eq:denoise_plus} as
\begin{equation}
    z_t = \frac{x_{t-1} - \hat{\mu}_t(x_t, x_{t+1})}{\sigma_t}, \qquad t=T,..,1    
\end{equation}
with $\hat{\mu}$ and $\sigma_t$ as defined above.
We base our implementation on the multistep variant of sde-dpm-solver++, which only requires one evaluation of the DM at each diffusion timestep by reusing the estimates from the previous step.
The number of timesteps can be reduced further by stopping the inversion at an intermediate step $t < T$ and starting the generation at that step. Empirically, we observed that $t \in [0.9T, 0.8T]$ usually produces edits of the same fidelity as $t=T$, supporting observations in previous work \cite{meng2022sdedit, hubermanspiegelglas2023edit} that earlier timesteps are less relevant to the edit.
\begin{figure*}[t]
    \centering
    \includegraphics[width=.8\linewidth]{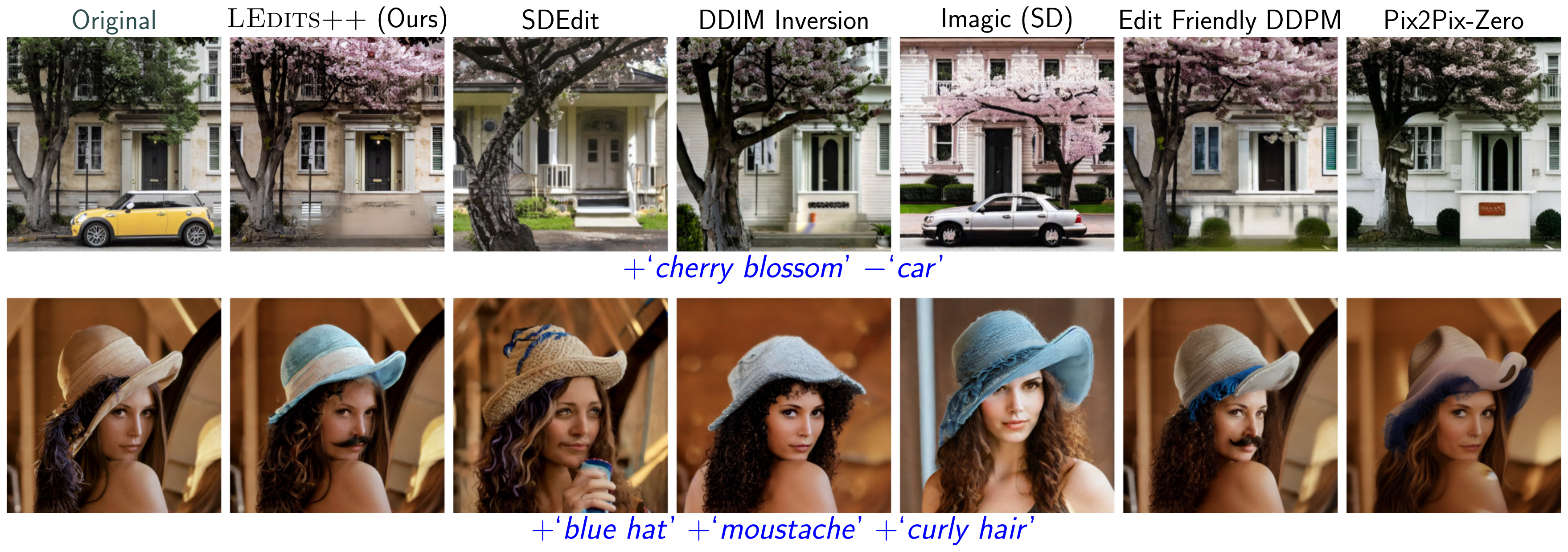}
    \vskip -0.5em
    \caption{Comparison of image editing methods. (top) \model~is the only method to restrict edits to the tree leaves and position of the car. (bottom) Ours~is the only approach faithfully executing all three edits and keeping changes minimal. (Best viewed in color)}
    \label{fig:qualitative_comparison}
\end{figure*}
\begin{table*}[t]
    \centering
    \small
    {\tabcolsep=8pt
    \begin{tabular}{l c c c c c}
        \multirow{2}{*}{Method} & \multirow{2}{*}{\makecell{Reconstruction\\Error (RMSE)}\,$\downarrow$}  & \multirow{2}{*}{\makecell{Execution\\Time (s)}\,$\downarrow$} & \multirow{2}{*}{\makecell{Variation/\\Sampling}} &\multirow{2}{*}{\makecell{Semantic\\Grounding}} &\multirow{2}{*}{\makecell{Multi-\\Editing}} \\ \\ \hline
        SDEdit \cite{meng2022sdedit}& 0.81 $_{\pm0.07}$ & 
        \phantom{00}2.10 $_{\pm0.02}$  & \checkmark & \ding{55} & \ding{55} \\        
        Imagic \cite{kawar2023imagic}& 0.58  $_{\pm0.12}$ & 
        349.98  $_{\pm0.45}$ & \checkmark & \ding{55} & \ding{55}\\
        
        Vanilla DDIM Inversion & 0.22 $_{\pm0.10}$ & 
        \phantom{0}37.23 $_{\pm0.04}$ & \ding{55} & \ding{55} & \ding{55}\\
        Pix2Pix-Zero \cite{parmar2023zeroshot}& 0.20 $_{\pm0.09}$ & \phantom{0}56.78 $_{\pm0.14}$ & (\checkmark) & \checkmark & \ding{55}\\
        
       DiffEdit  \cite{couarion2023diffedit} &  {$0.13_{\pm0.03}$} & \phantom{0}27.65 $_{\pm0.03}$ & \checkmark & \checkmark & \ding{55}\\
        Edit-friendly DDPM  \cite{hubermanspiegelglas2023edit} & \textbf{0.00} \phantom{$_{\pm0.05}$}& 
        \phantom{0}10.36 $_{\pm0.05}$& \checkmark & \ding{55} & \ding{55} \\
        \model~(Ours) & \textbf{0.00} \phantom{$_{\pm0.05}$}& 
        \phantom{0}\phantom{0}\textbf{1.78} $_{\pm0.03}$ & \checkmark & \checkmark  & \checkmark\\
    \end{tabular}}
    \vskip -0.5em
    \caption{Comparing key properties for diffusion-based image editing techniques, with \model~offering clear methodological benefits. Due to \model's efficient perfect inversion, it is the fastest and error-free method. At the same time, its methodology is the only enabling versatility in terms of \textit{variation}, \textit{semantic grounding}, and \textit{multi-editing}. Subscript numbers indicate standard deviation.
    }
    \label{tab:properties}
    \vskip -0.5em
\end{table*}

\paragraphneurips{Component 2: Textual Editing.}
After creating our reconstruction sequence $x_1,...,x_T$ and calculating the respective $z_t$, we now edit the image by manipulating the noise estimate $\hat{\epsilon}_\theta$ based on a set of edit instructions $\{e_i\}_{i \in I}$. 
We devise a dedicated guidance term for each concept $e_i$ based on conditioned and unconditioned estimates. Let us formally define \model's guidance by starting with a single editing prompt $e$. We compute
\begin{equation}\label{eq:sega}
    \hat{\epsilon}_\theta({x}_t, c_e) := \hat{\epsilon}_\theta({x}_t) + \gamma(x_t, c_e) 
\end{equation}
with guidance term $\gamma$. Consequently, setting $\gamma=0$ will reconstruct the input image $x_0$. We construct $\gamma$ to push the unconditioned score estimate $\hat{\epsilon}_\theta({x}_t)$---i.e. the input image reconstruction---away from/towards the edit concept estimate $\hat{\epsilon}_\theta({x}_t, c_e)$, depending on the guidance direction:
\begin{equation}
    \gamma(x_t, c_e) = \phi(\psi; s_e, \lambda) \psi(x_t, c_e)
    \label{eq:sem_guidance}
\end{equation}
where $\phi$ applies an edit guidance scale $s_e$ element-wise, and $\psi$ depends on the edit direction: $\psi(x_t, c_e)=$
\begin{equation}
    \begin{cases}
  \hat{\epsilon}_\theta({x}_t, c_e) - \hat{\epsilon}_\theta({x}_t)   \quad \text{if pos. guidance} \\
   -\big(\hat{\epsilon}_\theta({x}_t, c_e) - \hat{\epsilon}_\theta({x}_t)\big)  \quad  \text{if neg. guidance}
   \label{eq:guidance_direction}
\end{cases}
\end{equation}
Thus, changing the guidance direction is reflected by the direction between $\hat{\epsilon}_\theta({x}_t,\! c_e)$ and $\hat{\epsilon}_\theta({x}_t)$.
The term $\phi$ identifies those dimensions of the image and respective $\hat{\epsilon}_\theta$ that are relevant to a prompt $e$. Consequently, $\phi$ returns $0$ for all irrelevant dimensions and a scaling factor $s_e$ for the others. We describe the construction of $\phi$ in detail below. Larger $s_e$ will increase the effect of the edit, and $\lambda \in (0,1)$ reflects the percentage of the pixels selected as relevant by $\phi$. Notably, for a single concept $e$ and uniform $\phi=s_e$, Eq.~\ref{eq:sega} generalizes to the classifier-free guidance term in Eq.~\ref{eq:classifier_free}.

For multiple $e_i$, we calculate $\gamma_t^i$  as described above, with each defining their own hyperparameter values $\lambda^i$, $s_e^i$. The sum of all $\gamma_t^i$ results in 
\begin{equation}
     \hat{\gamma}_t(x_t, c_{e_i}) = \sum\nolimits_{i \in I} \gamma_t^i(x_t, c_{e_i}) 
    \label{eq:mult_sumup}
\end{equation}

\paragraphneurips{Component 3: Semantic Grounding.}
The masking term~$\phi$ (Eq.~\ref{eq:sem_guidance}) is the intersection (pointwise product) of binary masks $M^1$ and $M^2$ combined with scaling factor $s_e$:
\begin{equation}
    \phi(\psi; s_{e_i}, \lambda) = s_{e_i} M^1_i M^2_i
\end{equation}
where $M^1_i$ is a binary mask generated from the U-Net’s cross-attention layers and $M^2_i$ is a binary mask derived from the 
noise estimate. Intuitively, each mask is an importance map, where $M^1_i$ is more strongly grounded than $M^2_i$, but of significantly coarser granularity. Therefore, the intersection of the two yields a mask both focused on relevant image regions and of fine granularity. With \model,~we empirically demonstrate that these maps can also capture regions of an image relevant to an editing concept that is not already present. Specifically for multiple edits, calculating a dedicated mask for each edit prompt ensures that the corresponding guidance terms remain largely isolated, limiting interference between them.


Formally, at each time step $t$, a U-Net forward pass with editing prompt $e_i$ is performed to generate cross-attention maps for each token of the editing prompt. All cross-attention maps of the smallest resolution (e.g.,~$16\!\!\times\!\!16$ for SD) are averaged over all heads and layers, and the resulting maps are summed over all editing tokens, resulting in a single map $A_t^{e_i} \in R^{16\times16}$. Importantly, we utilize the same U-Net evaluation $\hat{\epsilon}_\theta({x}_t, c_e)$ already performed in Eq.~\ref{eq:guidance_direction} to produce $M^1$ with minimal overhead.
Each map $A_t^{e_i}$ is up-sampled to match the size of $x_t$. Cross-attention mask $M^1$ is derived by calculating the $\lambda$-th percentile of \mbox{up-sampled $A_t^{e_i}$} and
\begin{equation}
    \label{eq:maskm1}
    M^1_i = \begin{cases}
    1 \quad \text{if } |A_t^{e_i}| \geq \eta_\lambda(|A_t^{e_i}|) \\
    0 \quad \text{else}
    \end{cases}
\end{equation}
where $\eta_\lambda(|\!\cdot\!|)$ is the $\lambda$-th percentile.
By definition, $M^1$ only selects image regions that correlate strongly with the editing prompt, and $\lambda$ determines the size of this selected region.

The fine-grained mask $M^2$ is calculated based on the guidance vector $\psi$ of noise estimates derived in Eq.~\ref{eq:guidance_direction}. The difference between unconditioned and conditioned $\hat{\epsilon}_\theta$, generally captures outlines and object edges of $x_t$. Consequently, the largest absolute values of $\psi$ provide meaningful segmentation information of fine granularity for $M^2$
\begin{equation}
    \label{eq:maskm2}
    M^2 = \begin{cases}
    1 \quad \text{if } |\psi| \geq \eta_\lambda(|\psi|) \\
    0 \quad \text{else}
    \end{cases}
\end{equation}
In general, threshold $\lambda$ should correspond to the performed edit. Changes affecting the entire image, such as style transfer, should choose smaller $\lambda~(\rightarrow 0$), whereas edits targeting specific objects or regions should use $\lambda$ proportional to the region's prominence in the image.

\section{Properties of \model}\label{sec:properties}\noindent
With the fundamentals of \model~established, we next showcase its unique properties and capabilities. 

\paragraphneurips{Efficiency.}
First off, \model~offers substantial performance improvements over other image editing methods. In Tab.~\ref{tab:properties}, we provide a qualitative runtime comparison, with all methods being implemented for Stable Diffusion (SD) 1.5 \cite{rombach2022High}. As a parameter-free approach, \model~does not require any computationally expensive fine-tuning or optimization. Consequently, \model~is orders of magnitude faster than methods like Imagic \cite{kawar2023imagic} or Pix2Pix-Zero \cite{parmar2023zeroshot}. 
Further, we only need to invert the same number of diffusion steps used at inference, which results in significant runtime improvements over the standard DDIM inversion (21x). In addition to efficient inversion, we use a recent, fast scheduler that generally requires fewer total steps, further boosting performance. This way, \model~is six times faster than recent DDPM inversion \cite{hubermanspiegelglas2023edit} and on par with fast but poor-quality SDEdit \cite{meng2022sdedit}. 

\paragraphneurips{Versatility.}
In addition to its efficiency, \model~remains versatile, enabling sheer limitless editing possibilities. 
In Fig.~\ref{fig:basic_examples}, we showcase a broad range of edit types. \model~facilitates fine-grained edits (adding/removing glasses) and holistic changes such as style transfer (painting/sketch). Furthermore, object removal and replacement facilitate even more image editing tasks. Importantly, the overall image composition is preserved in all cases.
To our knowledge, \model~is the only diffusion-based image editing method inherently supporting multiple edits in isolation, which allows for more complex image manipulation. Fig.~\ref{fig:qualitative_comparison} highlights \model~benefits over previous methods. Our method produces the highest edit fidelity and is the only approach capable of faithfully executing multiple, simultaneous instructions. Moreover, \model~also makes the least changes to unrelated objects and the overall background and composition of the image. 

Lastly, the editing versatility benefits from the stochastic nature of the perfect but non-deterministic inversion. \model~provides meaningful image variations by resampling $\tilde{\epsilon}_t$. 
Additionally, the visual expression of each concept in the edited image scales monotonically with $s_e$, and the direction and magnitude of each concept can be varied freely. We present examples of both features in App.~\ref{app:properties}.
%
%

\begin{figure}
    \centering
    \includegraphics[width=0.8\linewidth]{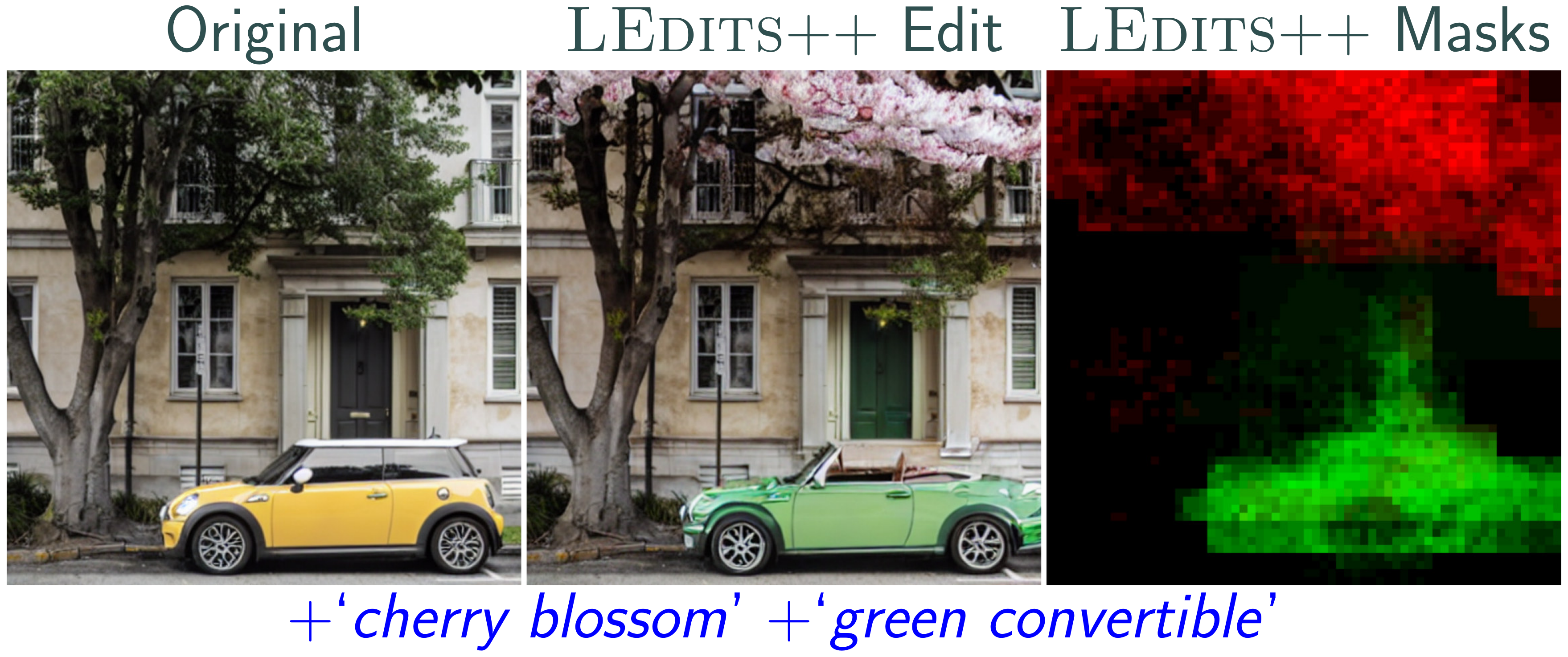}
    \vskip -0.5em
    \caption{Exemplary edit performed with \model~in only 25 diffusion steps with SD1.5. We apply a complex, compounded edit and ground each to a semantically reasonable image region.}
    \label{fig:ledits_masks}
    \vskip -1em
\end{figure}

\paragraphneurips{Precision.}
Furthermore, \model's methodology keeps edits concise and avoids unnecessary deviations from the input image (Fig.~\ref{fig:qualitative_comparison}). First, the perfect inversion will reconstruct the exact input image if no edit is applied (cf. Sec.~\ref{subsec:ledits}). Consequently, we already improve on faithfulness to the input image even before applying any edits. This benefit over other methods is highlighted by the reconstruction error in  Tab.~\ref{tab:properties}.
%
%
Second, implicit masking will semantically ground each edit to relevant image regions. This is specifically important for editing multiple concepts at the same time. While other methods only utilize one prompt for all instructions, \model~isolates edits from each other (Eq.~\ref{eq:mult_sumup}). Thus, we get dedicated masks for each concept as shown in Fig.~\ref{fig:ledits_masks}. This design ensures that each instruction (e.g.,~red mask for `cherry blossom') will be only applied where necessary. Subsequently, we provide further evidence for the efficacy of \model's masking approach.

\section{Semantically Grounded Image Editing}\label{sec:masking}\noindent
%
Cross-attention maps of DMs have been used extensively to ground regions of interest during image generation semantically \cite{hertz2022prompt,cao2023masactrl,chefer2023attend,parmar2023zeroshot}. Nonetheless, these have not been combined with noise-based masks so far and thus lack fine granularity. 
Hence, we empirically evaluate the quality of implicit masks, i.e., attention maps $M^1$ and noise maps $M^2$ 
(Eq.~\ref{eq:maskm1} and \ref{eq:maskm2}) in the \model~setup. We use a broad segmentation task for common objects as a proxy to measure the performance of implicit masks in identifying relevant image areas from edit instructions. Specifically, we utilize segmentation masks from the COCO panoptic segmentation challenge \cite{lin2014coco}. 
For each unique object in an image, we retrieve the masks $M^1, M^2$, and their intersection per diffusion step. We use the (semantic) class label (e.g.~`person' or `TV') as editing concept $e$. We consider masks at each of 50 total diffusion steps without actually editing the input image. Furthermore, we approximate mask threshold $\lambda$ based on the relative size of an object's bounding box.

\begin{figure}
    \centering
    \includegraphics[width=.9\linewidth]{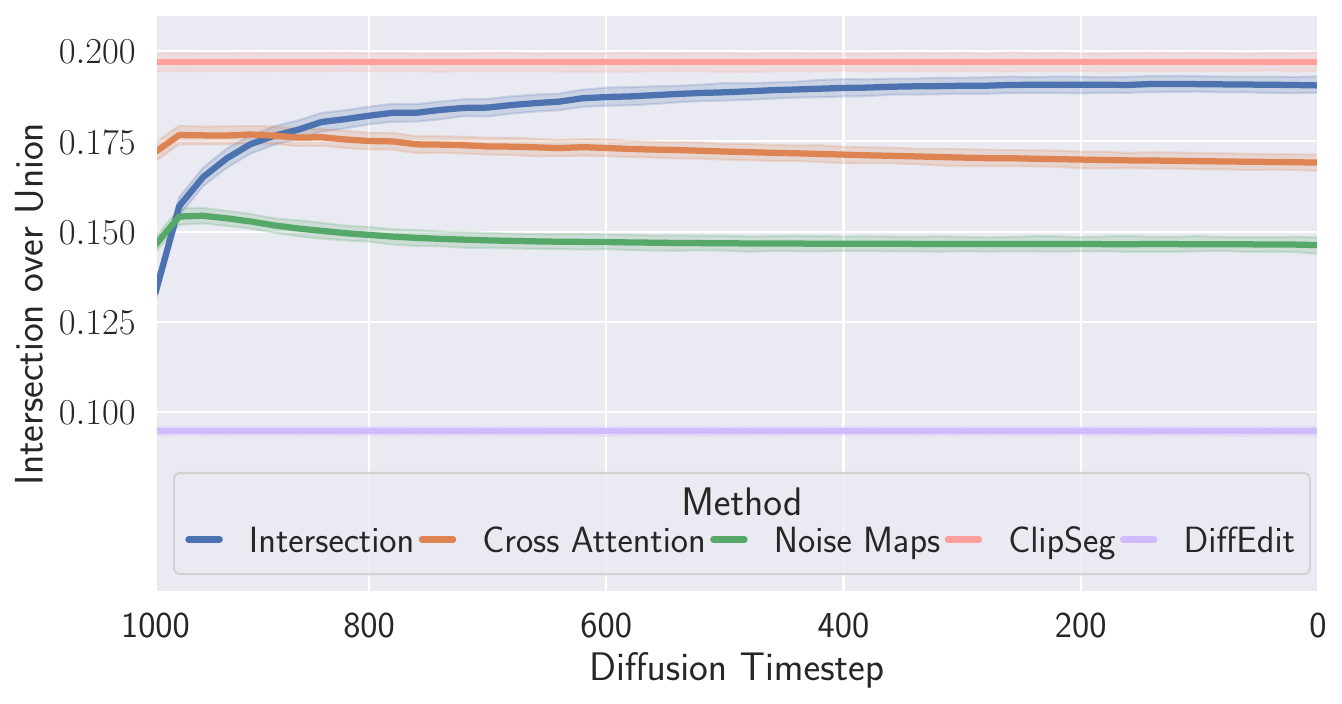}
    \vskip -0.5em
    \caption{Semantic segmentation quality of \model. We show the intersection over union (higher is better) for COCO panoptic segmentation. The intersection masks outperform each by a clear margin, close to the CLIPSeg reference. (Best viewed in color)}
    \label{fig:coco_seg}
    \vskip -1em
\end{figure}

\paragraphneurips{Concise masks with \model.}
Fig.~\ref{fig:coco_seg} shows 
implicit masking as a reliable means to identify relevant image regions. Importantly, the intersection of both cross-attention masks $M^1$ and noise maps $M^2$ clearly outperforms each separate mask. 
The overall performance is even similar to a dedicated CLIPSeg model \cite{lueddecke2022image}, despite \model~masks being implicitly calculated at inference with only minimal overhead. 
At the same time, \model's masking is superior to DiffEdit's \cite{balaji2022ediffi}.
Consequently, our method's intersection of cross-attention masks and noise maps provides strong semantic grounding while being efficient during image manipulation to ensure precise editing. 

\section{Image Editing Evaluation}\label{sec:experiments}\noindent
Let us now compare \model~to current SOTA methods for image manipulation on two benchmarks.

\subsection{Editing Multiple Concepts}\label{sec:mult_cond}\noindent
First, we investigate the complex task of performing multiple edits simultaneously. We rely on a well-established setup for semantic image manipulation \cite{brack2023Sega} to evaluate multi-conditioned attribute manipulation in facial images. In our experiment, we consider 100 images from the CelebA dataset \cite{liu2015deep}.
For each image, we simultaneously edit three attributes out of a set of five, leading to ten total combinations of edit concepts. Further, we perform each edit across ten different seeds, resulting in 10,000 evaluated images for each method and hyperparameter setting, over 1M images in total.
As measures for comparison, we employ CLIP and LPIPS scores. CLIP measures the text-to-image similarity of the edit instruction to the edited image, and LPIPS measures the image-to-image similarity of the real to the edited image.
This way, we assess the trade-off between the versatility of edits (CLIP) and the precision of those manipulations (LPIPS). 
We implement all methods based on SD1.5 and provide more details in App.~\ref{app:expdetails}.

\begin{figure}
    \centering
    \includegraphics[width=.9\linewidth]{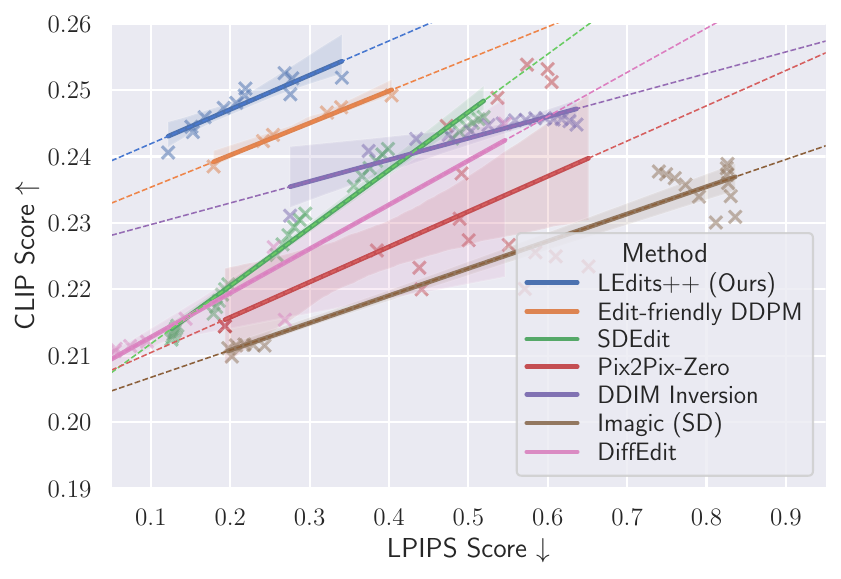}
    \vskip -1em
    \caption{Comparison of instruction-alignment vs.~image similarity trade-off for different editing methods. Results were reported for simultaneous manipulation of three facial attributes on CelebA. We plot CLIP scores (higher is better) of the target attributes against LPIPS similarity (lower is better). \model~clearly outperforms all competing methods.
     (Best viewed in color)}
    \label{fig:celeb_a_comp}
    \vskip -1em
\end{figure}

\paragraphneurips{\model~outperforms competing methods.}
Fig.~\ref{fig:celeb_a_comp} shows the resulting CLIP vs.~LPIPS plots for all methods. The top left corner represents the ideal editing method with maximum edit alignment without deviating from the initial image. Generally, one can observe a natural trade-off between versatility and precision for all methods, i.e.,~higher image-to-text alignment comes at the expense of lower similarity to the original image. \model~is closest to the ideal region and thus clearly outperforms the other methods. 
In particular, the outputs remain close to the original image (low LPIPS scores), thanks to the precise implicit masking. At the same time, it faithfully performs the edits (high CLIP scores) 
due to the dedicated, isolated editing for each concept.
The depicted scores reflect our qualitative inspections for Pix2Pix-Zero and Imagic on such complex manipulations (cf.~Fig.~\ref{fig:qualitative_comparison}). We observed that these methods often break---either failing to perform all three edits and/or drastically altering the input image. Only edit-friendly DDPM \cite{hubermanspiegelglas2023edit} and \model~reliably achieve the maximum average CLIP score of over $0.25$. This value seems to represent an upper bound according to our manual investigations, as each attribute is edited correctly for all input images, and higher scores are not observed. 
Despite being computationally very efficient, \model~faithfully executes each edit instruction while keeping the changes to the input low, highlighting the method's versatility and precision.

\subsection{TEdBench(++)}
\begin{figure*}
    \centering
    \begin{subfigure}[b]{.55\textwidth}
         \centering
          \includegraphics[height=16em]{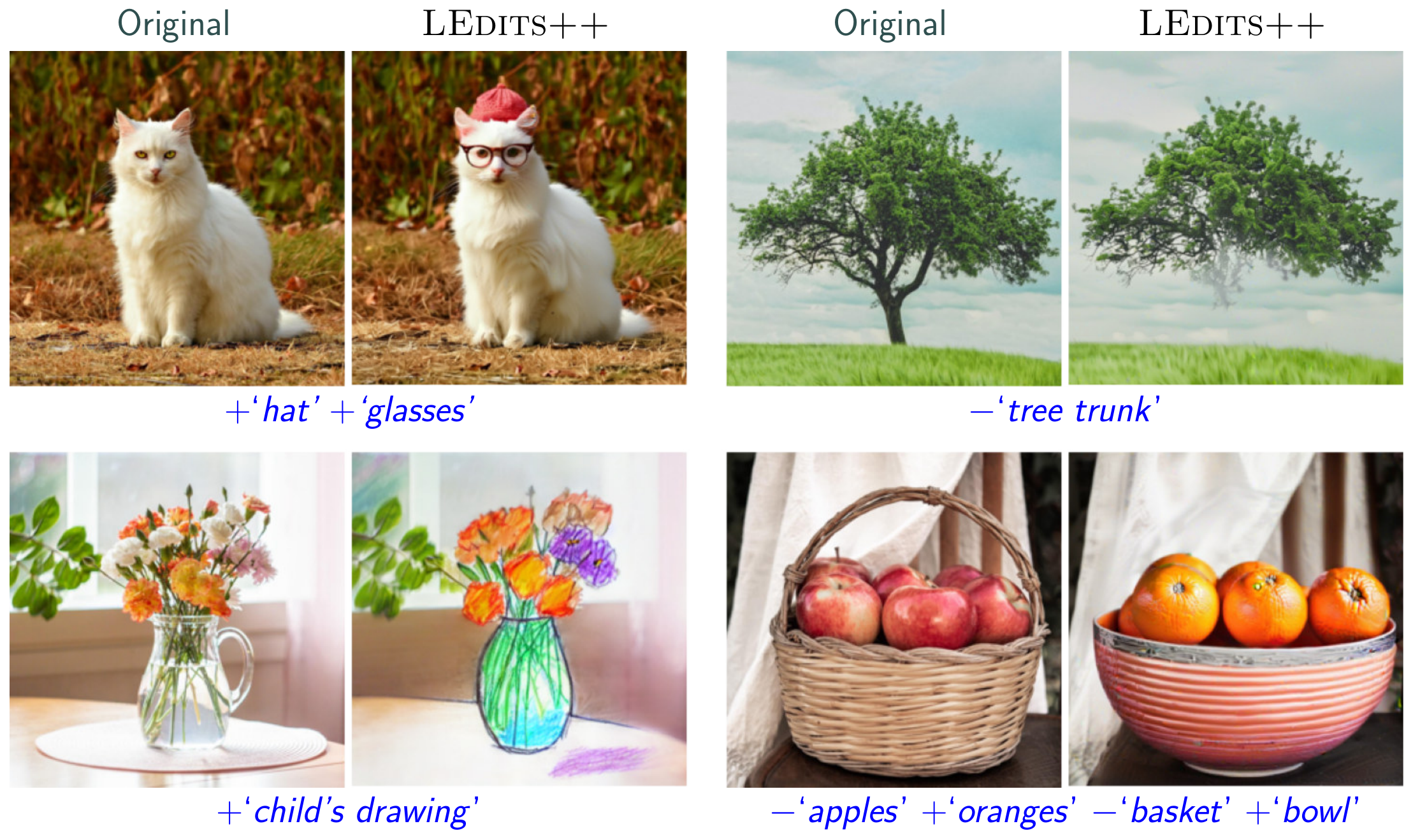}
         \caption{Novel challenging examples of TEdBench++ and \model~applied, showcasing the versatility of supported edits. }
         \label{fig:tedbench++}
     \end{subfigure}
     \hfill
     \begin{subfigure}[b]{.40\textwidth}
         \centering
         \includegraphics[height=16em]{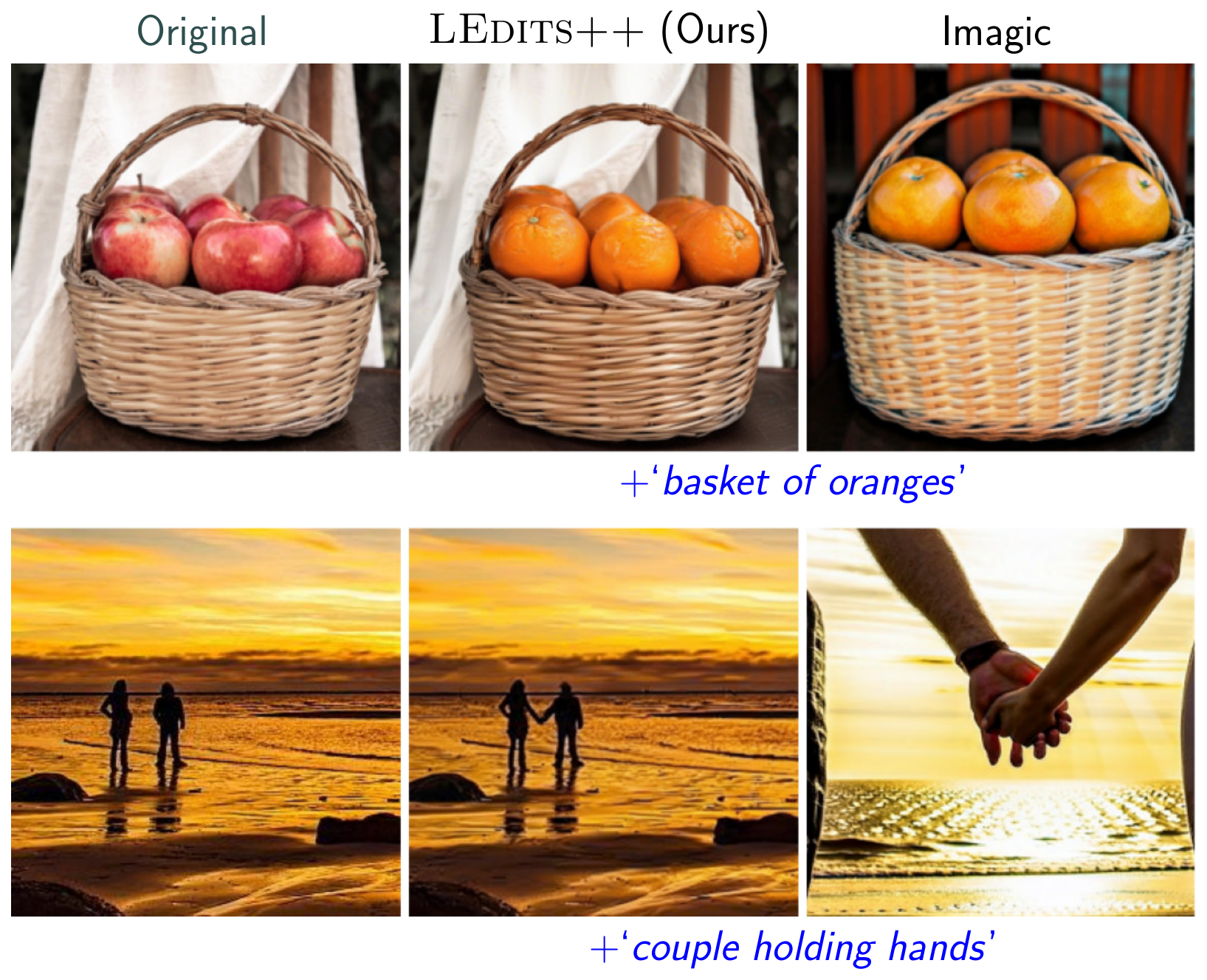}
         \caption{Qualitative comparison of \model~and Imagic on TEdBench, clearly highlighting the performance improvement.}
         \label{fig:tedbench_comparison}
     \end{subfigure}
    \vskip -0.5em
     
    \caption{Benchmark examples for \model~and Imagic on TEdBench(++). (Best viewed in color)}
    \vskip -1em
    \label{fig:creative_examples}
\end{figure*}

\noindent
%
%
Next, we
investigate the versatility of \model's editing capabilities by running the Textual Editing Benchmark (TEdBench \cite{kawar2023imagic}), a collection of 100 input images paired with textual edit instructions. However, we observed a variety of inconsistencies in TEdBench and a lack of relevant editing tasks. Therefore, we propose TEdBench++ (Fig.~\ref{fig:tedbench++} and App.~\ref{app:tedbench++}), a more challenging revised benchmark now containing 120 entries in total.\footnote{\tiny \url{https://huggingface.co/datasets/AIML-TUDA/TEdBench_plusplus}}
%
We addressed misspellings and rephrased ambiguous and inconclusive instructions. 
In addition to resolving these issues, we added instructions targeting challenging types of image manipulations previously not included in TEdBench: multi-conditioning, object/concept removal, style transfer, and complex replacements (Fig.~\ref{fig:tedbench++}). We provide more details in App.~\ref{app:tedbench++}.

\begin{table}
    \centering
    \small
    {
    \tabcolsep=4.1pt
    \begin{tabular}{c|rlrl}
         \multicolumn{1}{c}{} & \multicolumn{2}{c}{TEdBench} & \multicolumn{2}{c}{TEdBench++} \\
         & SR $\uparrow$ & LPIPS $\downarrow$ & SR $\uparrow$ & LPIPS $\downarrow$ \\
         \hline         
         Imagic w/ SD1.5&0.55 &  \phantom{.}0.56 & 0.58 &  \phantom{.}0.57\\
         \model~w/ SD1.5 & \textbf{0.75} &  \phantom{.}\textbf{0.28} & \textbf{0.79} &  \phantom{.}\textbf{0.30}\\
         \hline
         Imagic w/ Imagen \cite{kawar2023imagic} & 0.83 &  \phantom{.}0.59 & ---\phantom{.} &  \phantom{..}--- \\
         \model~w/ SD-XL & \textbf{0.84} &  \phantom{.}\textbf{0.33} & \textbf{0.87} &  \phantom{.}\textbf{0.34}
    \end{tabular}
    }
    \vskip -0.5em
    \caption{Success rate (SR) and LPIPS scores on the original TedBench \cite{kawar2023imagic} and our revised version (TEdBench++). We compare Imagic to \model~based on different DMs and find the latter to outperform on both metrics and benchmarks.}
    \label{tab:tedbench}
    \vskip -1.5em
\end{table}

We compare \model~on TEdBench(++) to one of the strongest editing methods, Imagic \cite{kawar2023imagic} with Imagen \cite{saharia2022photorealistic}. Since both
are not publicly available, we can only compare to this specific combination of DM and editing method using Kawar~\textit{et al.'s}~\cite{kawar2023imagic}~curated outputs for TEdBench. Additionally, we, therefore, cannot combine \model~with Imagen\cite{saharia2022photorealistic} and instead use a similarly advanced diffusion model, SD-XL \cite{podell2023sdxl}. However, to not only compete for the best fidelity outputs but focus the evaluation on methodological differences---not the pre-trained DM---we also compare both methods implemented with SD1.5.
We provide further details in App.~\ref{app:expdetails}.
%

\paragraphneurips{\model~edits images reliably.}
We first asked users to assess the overall success of edits, i.e., if an edit instruction was faithfully realized for a given input image. 
The results in Tab.~\ref{tab:tedbench} show that \model~outperforms Imagic on TEdBench despite a greatly reduced runtime (Tab.~\ref{tab:properties}). The difference is even stronger when comparing both methods on the same pre-trained DM, i.e.,~SD1.5. The high success rate on TEdBench++ (87\%) and the examples shown in Fig.~\ref{fig:basic_examples} and~\ref{fig:tedbench++} once again highlight \model's versatility. 
Overall, our proposed method can reliably perform a diverse set of editing instructions for real images. 

\paragraphneurips{High-quality edits with \model.}
While investigating both methods' performance
we observed a substantial difference in edit quality. The examples in Fig.~\ref{fig:tedbench_comparison} particularly highlight the discrepancy in compositional robustness and object coherence. Hence, we also assessed both methods' editing quality on TEdBench(++). 
We focus on samples where both methods performed a successful edit, i.e., were labeled as successful by users.
We show the perceptual similarity (LPIPS) to the input image in Tab.~\ref{tab:tedbench}. One can observe that the LPIPS scores for \model~are much lower than for Imagic, empirically supporting the qualitative examples in Fig.~\ref{fig:tedbench_comparison}. 
When manually inspecting the generated images, we often found Imagic to generate a completely \textit{new} image based on the edit instruction, entirely disregarding the input image (cf.~App.~Fig.~\ref{fig:failure2a}). 




\section{Discussion} \label{sec:discussion} \noindent
Let us now discuss open research questions and limitations.


\paragraphneurips{Model Dependency.}
While \model~achieves impressive results on a large variety of image manipulation tasks, there are external factors to consider. 
Since the method is architecture-agnostic, it can be easily used with any DM.
At the same time, the general editing quality strongly depends on the overall capabilities of the underlying pre-trained DM. Naturally, more capable models will also enable better edits. But, at times, specific editing instructions may fail because the used DM does not have a decent representation of the targeted concept to begin with. One example is the model failing to edit a giraffe to be \textit{sitting} since the underlying DM generally fails to generate this pose (cf.~App.~\ref{app:limitations}).
This effect can also clearly be seen in Tab.~\ref{tab:tedbench}, with the editing success rate of a method varying strongly between DMs. Although the same image editing method is employed (\model), the more capable SD-XL variant outperforms the weaker SD1.5 model. Nonetheless, this means that the architecture-agnostic \model~will benefit from increasingly powerful DMs. 

\paragraphneurips{Coherence Trade-offs.}
Next to the benefits of \model's semantic grounding, there are also downsides to this approach. Overall, implicit masking limits changes to relevant portions of the image and achieves strong coherence with the original image composition. 
Yet, the object and its identity within the masked area may change based on 
various factors. Generic prompts, like ``a standing cat'' (cf.\space App.~\ref{app:limitations}), do not contain detailed information about this specific object (``cat''). Thus, an edit with this prompt does not guarantee to preserve object identity, particularly for strong hyperparameters. We observed that fine-tuning approaches like Imagic make the opposite trade-off, better preserving the object identity while changing the background and image composition substantially (cf. App.~\ref{app:limitations}). A potential remedy for a loss in object coherence with \model, is more descriptive edit prompting, e.g.~using textual inversion \cite{gal2022textual}.

Lastly, the automatically-inferred implicit masks allow for easy use of \model~without users tediously providing masks. Nonetheless, user intentions are diverse and cannot always be automatically inferred.
Sometimes, individual user masks provide better control over the editing process. Such user masks can be easily integrated into \model~(cf.~App.~\ref{app:limitations}), wherefore we encourage future research in this promising direction. 


\paragraphneurips{Societal Impact.}
\model~is an easy-to-use image editing technique that lowers the barrier for users and puts them in control for fruitful human-machine collaboration.
Yet, the underlying text-to-image models offer both promise and peril, as highlighted by prior research~\cite{bianchi2023easily,friedrich2023fair}.
The (societal) biases within these models will also impact image editing applications \cite{friedrich2023fair}. Moreover, image manipulation can also be used adversarially to generate inappropriate \cite{schramowski2022safe} or fake content. Hence, we advocate for a cautious deployment of generative models together with image editing methods.

%

\section{Conclusion} \noindent
We introduced \model, an efficient yet versatile and precise method for textual image manipulation with diffusion models. It facilitates the editing of complex concepts in real images. Our approach requires no finetuning nor optimization, can be computed extremely efficiently, and is architecture agnostic. At the same time, it perfectly reconstructs an input image and uses implicit masking to limit changes to relevant image regions, thus editing precisely. Our large experimental evaluation confirms the efficiency, versatility, and precision of \model~and its components, as well as its benefits over several related methods. 

\vspace{-0.5em}
\paragraph{Acknowledgements}
We gratefully acknowledge support from the BMBF (Grant No 01IS22091). This work benefited from the ICT-48 Network of AI Research Excellence Center ``TAILOR'' (EU Horizon 2020, GA No 952215), the Hessian research priority program LOEWE within the project WhiteBox, the HMWK cluster projects ``Adaptive Mind'' and ``Third Wave of AI'', and from the NHR4CES.

\newpage
{\small
\bibliographystyle{ieee_fullname}
\bibliography{bibliography}

\begin{thebibliography}{10}\itemsep=-1pt

\bibitem{avrahami2023break}
Omri Avrahami, Kfir Aberman, Ohad Fried, Daniel Cohen-Or, and Dani Lischinski.
\newblock Break-a-scene: Extracting multiple concepts from a single image.
\newblock {\em SIGGRAPH Asia}, 2023.

\bibitem{avrahmi2022blended}
Omri Avrahami, Ohad Fried, and Dani Lischinski.
\newblock Blended latent diffusion.
\newblock {\em arXiv:2206.02779}, 2022.

\bibitem{balaji2022ediffi}
Yogesh Balaji, Seungjun Nah, Xun Huang, Arash Vahdat, Jiaming Song, Karsten
  Kreis, Miika Aittala, Timo Aila, Samuli Laine, Bryan Catanzaro, Tero Karras,
  and Ming{-}Yu Liu.
\newblock {eDiff-I}: Text-to-image diffusion models with an ensemble of expert
  denoisers.
\newblock {\em arXiv:2211.01324}, 2022.

\bibitem{bianchi2023easily}
Federico Bianchi, Pratyusha Kalluri, Esin Durmus, Faisal Ladhak, Myra Cheng,
  Debora Nozza, Tatsunori Hashimoto, Dan Jurafsky, James Zou, and Aylin
  Caliskan.
\newblock Easily {Accessible} {Text}-to-{Image} {Generation} {Amplifies}
  {Demographic} {Stereotypes} at {Large} {Scale}.
\newblock In {\em Proceedings of {ACM} Conference on Fairness, Accountability,
  and Transparency ({FAccT})}, 2023.

\bibitem{brack2023Sega}
Manuel Brack, Felix Friedrich, Dominik Hintersdorf, Lukas Struppek, Patrick
  Schramowski, and Kristian Kersting.
\newblock Sega: Instructing text-to-image models using semantic guidance.
\newblock In {\em Proceedings of the Advances in Neural Information Processing
  Systems: Annual Conference on Neural Information Processing Systems
  ({NeurIPS})}, 2023.

\bibitem{brooks2022instructpix2pix}
Tim Brooks, Aleksander Holynski, and Alexei~A. Efros.
\newblock Instructpix2pix: Learning to follow image editing instructions.
\newblock In {\em Proceedings of the {IEEE/CVF} Conference on Computer Vision
  and Pattern Recognition ({CVPR})}, 2023.

\bibitem{cao2023masactrl}
Mingdeng Cao, Xintao Wang, Zhongang Qi, Ying Shan, Xiaohu Qie, and Yinqiang
  Zheng.
\newblock Masactrl: Tuning-free mutual self-attention control for consistent
  image synthesis and editing.
\newblock {\em arXiv:2304.08465}, 2023.

\bibitem{chefer2023attend}
Hila Chefer, Yuval Alaluf, Yael Vinker, Lior Wolf, and Daniel Cohen{-}Or.
\newblock Attend-and-excite: Attention-based semantic guidance for
  text-to-image diffusion models.
\newblock {\em {ACM} Trans. Graph.}, 42, 2023.

\bibitem{couarion2023diffedit}
Guillaume Couairon, Jakob Verbeek, Holger Schwenk, and Matthieu Cord.
\newblock Diffedit: Diffusion-based semantic image editing with mask guidance.
\newblock In {\em Proceedings of the International Conference on Learning
  Representations ({ICLR})}, 2023.

\bibitem{desimone2023whatis}
Zoe De~Simone, Angie Boggust, Arvind Satyanarayan, and Ashia Wilson.
\newblock What is a {Fair} {Diffusion} {Model}? {Designing} {Generative}
  {Text}-{To}-{Image} {Models} to {Incorporate} {Various} {Worldviews}.
\newblock {\em arXiv preprint arXiv:2309.09944}, 2023.

\bibitem{friedrich2023fair}
Felix Friedrich, Manuel Brack, Lukas Struppek, Dominik Hintersdorf, Patrick
  Schramowski, Sasha Luccioni, and Kristian Kersting.
\newblock Fair {Diffusion}: {Instructing} {Text}-to-{Image} {Generation}
  {Models} on {Fairness}.
\newblock {\em arXiv preprint arXiv:2302.10893}, 2023.

\bibitem{gal2022textual}
Rinon Gal, Yuval Alaluf, Yuval Atzmon, Or Patashnik, Amit~H. Bermano, Gal
  Chechik, and Daniel Cohen-Or.
\newblock An image is worth one word: Personalizing text-to-image generation
  using textual inversion.
\newblock {\em arXiv preprint arXiv:2208.01618}, 2022.

\bibitem{gandikota2023erasing}
Rohit Gandikota, Joanna Materzynska, Jaden Fiotto-Kaufman, and David Bau.
\newblock Erasing concepts from diffusion models.
\newblock In {\em Proceedings of the {IEEE/CVF} International Conference on
  Computer Vision ({ICCV})}, 2023.

\bibitem{hertz2022prompt}
Amir Hertz, Ron Mokady, Jay Tenenbaum, Kfir Aberman, Yael Pritch, and Daniel
  Cohen{-}Or.
\newblock Prompt-to-prompt image editing with cross attention control.
\newblock In {\em Proceedings of the International Conference on Learning
  Representations ({ICLR})}, 2023.

\bibitem{ho2020denoising}
Jonathan Ho, Ajay Jain, and Pieter Abbeel.
\newblock Denoising diffusion probabilistic models.
\newblock In Hugo Larochelle, Marc'Aurelio Ranzato, Raia Hadsell,
  Maria{-}Florina Balcan, and Hsuan{-}Tien Lin, editors, {\em Proceedings of
  the Advances in Neural Information Processing Systems: Annual Conference on
  Neural Information Processing Systems ({NeurIPS})}, 2020.

\bibitem{ho2022classifier}
Jonathan Ho and Tim Salimans.
\newblock Classifier-free diffusion guidance.
\newblock {\em arXiv:2207.12598}, 2022.

\bibitem{hubermanspiegelglas2023edit}
Inbar Huberman-Spiegelglas, Vladimir Kulikov, and Tomer Michaeli.
\newblock An edit friendly ddpm noise space: Inversion and manipulations.
\newblock {\em arXiv preprint arXiv:2304.06140}, 2023.

\bibitem{kawar2023imagic}
Bahjat Kawar, Shiran Zada, Oran Lang, Omer Tov, Huiwen Chang, Tali Dekel, Inbar
  Mosseri, and Michal Irani.
\newblock Imagic: Text-based real image editing with diffusion models.
\newblock In {\em Proceedings of the {IEEE/CVF} Conference on Computer Vision
  and Pattern Recognition ({CVPR})}, 2023.

\bibitem{lin2014coco}
Tsung-Yi Lin, Michael Maire, Serge Belongie, James Hays, Pietro Perona, Deva
  Ramanan, Piotr Doll{\'a}r, and C.~Lawrence Zitnick.
\newblock Microsoft coco: Common objects in context.
\newblock In {\em Proceedings of European Conference on Computer Vision
  ({ECCV})}, 2014.

\bibitem{liu2022Compositional}
Nan Liu, Shuang Li, Yilun Du, Antonio Torralba, and Joshua~B. Tenenbaum.
\newblock Compositional visual generation with composable diffusion models.
\newblock In {\em Proceedings of European Conference on Computer Vision
  ({ECCV})}, 2022.

\bibitem{liu2015deep}
Ziwei Liu, Ping Luo, Xiaogang Wang, and Xiaoou Tang.
\newblock Deep learning face attributes in the wild.
\newblock In {\em Proceedings of the {IEEE/CVF} International Conference on
  Computer Vision ({ICCV})}, December 2015.

\bibitem{lu2022dpmsolver}
Cheng Lu, Yuhao Zhou, Fan Bao, Jianfei Chen, Chongxuan Li, and Jun Zhu.
\newblock Dpm-solver: {A} fast {ODE} solver for diffusion probabilistic model
  sampling in around 10 steps.
\newblock In {\em NeurIPS}, 2022.

\bibitem{lu2023dpmsolver}
Cheng Lu, Yuhao Zhou, Fan Bao, Jianfei Chen, Chongxuan Li, and Jun Zhu.
\newblock Dpm-solver++: Fast solver for guided sampling of diffusion
  probabilistic models.
\newblock {\em arXiv preprint arXiv:2211.01095}, 2022.

\bibitem{lueddecke2022image}
Timo L\"uddecke and Alexander Ecker.
\newblock Image segmentation using text and image prompts.
\newblock In {\em Proceedings of the {IEEE/CVF} Conference on Computer Vision
  and Pattern Recognition ({CVPR})}, 2022.

\bibitem{luo2023latent}
Simian Luo, Yiqin Tan, Longbo Huang, Jian Li, and Hang Zhao.
\newblock Latent consistency models: Synthesizing high-resolution images with
  few-step inference.
\newblock {\em arXiv:2310.04378}, 2023.

\bibitem{meng2022sdedit}
Chenlin Meng, Yutong He, Yang Song, Jiaming Song, Jiajun Wu, Jun{-}Yan Zhu, and
  Stefano Ermon.
\newblock Sdedit: Guided image synthesis and editing with stochastic
  differential equations.
\newblock In {\em Proceedings of the International Conference on Learning
  Representations ({ICLR})}, 2022.

\bibitem{meng2023ondistillation}
Chenlin Meng, Robin Rombach, Ruiqi Gao, Diederik~P. Kingma, Stefano Ermon,
  Jonathan Ho, and Tim Salimans.
\newblock On distillation of guided diffusion models.
\newblock In {\em Proceedings of the {IEEE/CVF} Conference on Computer Vision
  and Pattern Recognition ({CVPR})}, 2023.

\bibitem{mokady2023nulltext}
Ron Mokady, Amir Hertz, Kfir Aberman, Yael Pritch, and Daniel Cohen{-}Or.
\newblock Null-text inversion for editing real images using guided diffusion
  models.
\newblock In {\em Proceedings of the {IEEE/CVF} Conference on Computer Vision
  and Pattern Recognition ({CVPR})}, 2023.

\bibitem{nichol2022glide}
Alexander~Quinn Nichol, Prafulla Dhariwal, Aditya Ramesh, Pranav Shyam, Pamela
  Mishkin, Bob McGrew, Ilya Sutskever, and Mark Chen.
\newblock {GLIDE:} towards photorealistic image generation and editing with
  text-guided diffusion models.
\newblock In {\em Proceedings of the International Conference on Machine
  Learning ({ICML})}. {PMLR}, 2022.

\bibitem{parmar2023zeroshot}
Gaurav Parmar, Krishna~Kumar Singh, Richard Zhang, Yijun Li, Jingwan Lu, and
  Jun{-}Yan Zhu.
\newblock Zero-shot image-to-image translation.
\newblock In {\em SIGGRAPH}, 2023.

\bibitem{pernias2023wuerstchen}
Pablo Pernias, Dominic Rampas, Mats~L. Richter, Christopher~J. Pal, and Marc
  Aubreville.
\newblock Wuerstchen: An efficient architecture for large-scale text-to-image
  diffusion models.
\newblock {\em arXiv:2306.00637}, 2023.

\bibitem{podell2023sdxl}
Dustin Podell, Zion English, Kyle Lacey, Andreas Blattmann, Tim Dockhorn, Jonas
  Müller, Joe Penna, and Robin Rombach.
\newblock Sdxl: Improving latent diffusion models for high-resolution image
  synthesis.
\newblock {\em arXiv:2307.01952}, 2023.

\bibitem{ramesh2022hierarchical}
Aditya Ramesh, Prafulla Dhariwal, Alex Nichol, Casey Chu, and Mark Chen.
\newblock Hierarchical text-conditional image generation with {CLIP} latents.
\newblock {\em arXiv preprint arXiv:2204.06125}, 2022.

\bibitem{rombach2022High}
Robin Rombach, Andreas Blattmann, Dominik Lorenz, Patrick Esser, and
  Bj{\"{o}}rn Ommer.
\newblock High-resolution image synthesis with latent diffusion models.
\newblock In {\em Proceedings of the {IEEE/CVF} Conference on Computer Vision
  and Pattern Recognition ({CVPR})}, 2022.

\bibitem{ronneberger2015unet}
Olaf Ronneberger, Philipp Fischer, and Thomas Brox.
\newblock U-net: Convolutional networks for biomedical image segmentation.
\newblock In {\em Proceedings of Medical Image Computing and Computer-Assisted
  Intervention ({MICCAI})}, pages 234--241, 2015.

\bibitem{rosenberg2023unbiased}
Harrison Rosenberg, Shimaa Ahmed, Guruprasad~V. Ramesh, Ramya~Korlakai Vinayak,
  and Kassem Fawaz.
\newblock Unbiased {Face} {Synthesis} {With} {Diffusion} {Models}: {Are} {We}
  {There} {Yet}?
\newblock {\em arXiv preprint arXiv:2309.07277}, 2023.

\bibitem{saharia2022photorealistic}
Chitwan Saharia, William Chan, Saurabh Saxena, Lala Li, Jay Whang, Emily
  Denton, Seyed Kamyar~Seyed Ghasemipour, Burcu~Karagol Ayan, S.~Sara Mahdavi,
  Rapha~Gontijo Lopes, Tim Salimans, Jonathan Ho, David~J. Fleet, and Mohammad
  Norouzi.
\newblock Photorealistic text-to-image diffusion models with deep language
  understanding.
\newblock {\em arXiv:2205.11487}, 2022.

\bibitem{samuel2023generating}
Dvir Samuel, Rami Ben-Ari, Simon Raviv, Nir Darshan, and Gal Chechik.
\newblock Generating images of rare concepts using pre-trained diffusion
  models.
\newblock {\em arXiv:2304.14530}, 2023.

\bibitem{schramowski2022safe}
Patrick Schramowski, Manuel Brack, Björn Deiseroth, and Kristian Kersting.
\newblock Safe latent diffusion: Mitigating inappropriate degeneration in
  diffusion models.
\newblock In {\em Proceedings of the {IEEE/CVF} Conference on Computer Vision
  and Pattern Recognition ({CVPR})}, 2023.

\bibitem{schuhmann2022laion}
Christoph Schuhmann, Romain Beaumont, Richard Vencu, Cade~W Gordon, Ross
  Wightman, Theo Coombes, Aarush Katta, Clayton Mullis, Mitchell Wortsman,
  Patrick Schramowski, Srivatsa~R Kundurthy, Katherine Crowson, Ludwig Schmidt,
  Robert Kaczmarczyk, and Jenia Jitsev.
\newblock Laion-5b: An open large-scale dataset for training next generation
  image-text models.
\newblock In {\em Proceedings of NeurIPS Datasets and Benchmarks}, 2022.

\bibitem{schwarz1999how}
Norbert Schwarz.
\newblock How the questions shape the answers.
\newblock {\em American Psychologist}, 1999.

\bibitem{song2021denoising}
Jiaming Song, Chenlin Meng, and Stefano Ermon.
\newblock Denoising diffusion implicit models.
\newblock In {\em Proceedings of the International Conference on Learning
  Representations ({ICLR})}, 2021.

\bibitem{Song2021score}
Yang Song, Jascha Sohl{-}Dickstein, Diederik~P. Kingma, Abhishek Kumar, Stefano
  Ermon, and Ben Poole.
\newblock Score-based generative modeling through stochastic differential
  equations.
\newblock In {\em Proceedings of the International Conference on Learning
  Representations ({ICLR})}, 2021.

\bibitem{valevski2022UniTune}
Dani Valevski, Matan Kalman, Yossi Matias, and Yaniv Leviathan.
\newblock Unitune: Text-driven image editing by fine tuning an image generation
  model on a single image.
\newblock {\em arXiv:2210.09477}, 2022.

\bibitem{wu2022uncovering}
Qiucheng Wu, Yujian Liu, Handong Zhao, Ajinkya Kale, Trung Bui, Tong Yu, Zhe
  Lin, Yang Zhang, and Shiyu Chang.
\newblock Uncovering the disentanglement capability in text-to-image diffusion
  models.
\newblock {\em arXiv:1212.08698}, 2022.

\bibitem{zhao2023unipc}
Wenliang Zhao, Lujia Bai, Yongming Rao, Jie Zhou, and Jiwen Lu.
\newblock Unipc: {A} unified predictor-corrector framework for fast sampling of
  diffusion models.
\newblock {\em arXiv:2302.04867}, 2023.

\end{thebibliography}
}





\begin{figure*}
\centering
 \includegraphics[width=.9\linewidth]{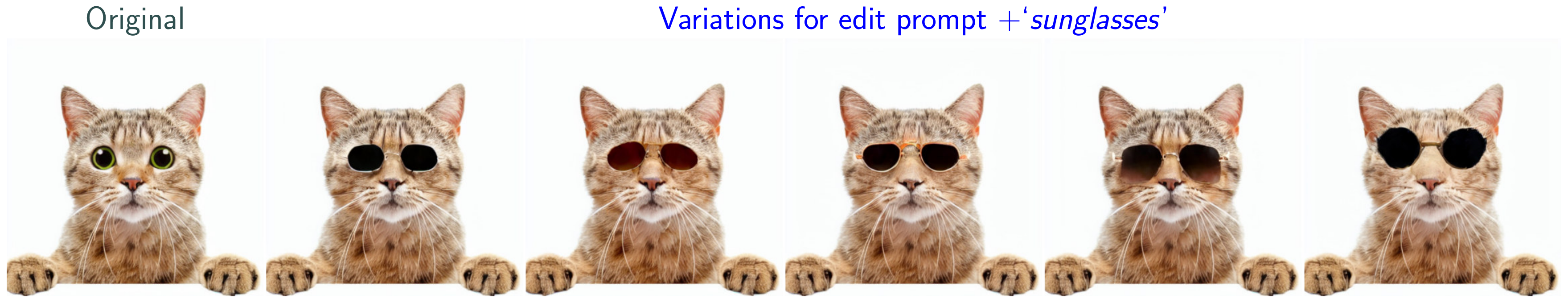}
 \caption{\model~easily produces variations of an edit (different (styles of) sunglasses) by resampling the inversion process. (Best viewed in color)}
 \label{fig:app_variations}
\end{figure*}
\begin{figure*}
   \centering
 \includegraphics[width=.9\linewidth]{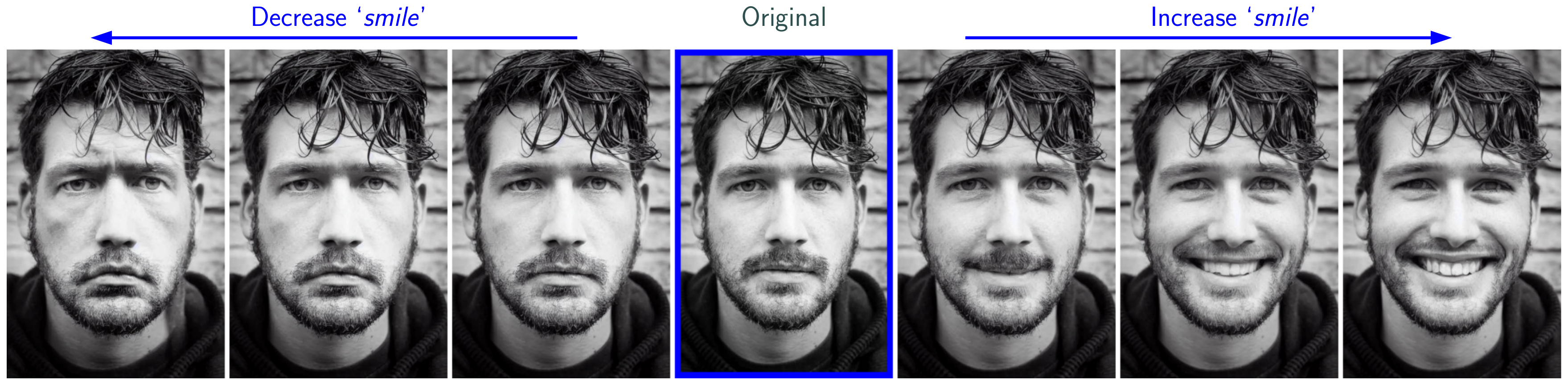}
 \caption{Monotonicity of editing scale with \model. The original image (middle) is edited with varying scales of the same edit ('smile'). The scale for ‘smile’ is semantically reflected in the images. (Best viewed in color)}
 \label{fig:app_monotonicity}
\end{figure*}


\newpage

\appendix

\section*{Appendix}
\section{Broader (Societal) Impact}

\noindent With \model, we aim to provide an easy-to-use image editing framework. It lowers the barrier of entry for experienced artists and novices alike, allowing them to unlock the full potential of generative AI in the pursuit of creative expression. Moreover, it puts the user in control for fruitful human-machine collaboration.
Crucially, current text-to-image models~\cite{ramesh2022hierarchical, nichol2022glide, saharia2022photorealistic} hold the potential to wield a profound influence on society. When applied in creative and design domains, their dual use offers both promise and peril, as highlighted by prior research~\cite{bianchi2023easily,friedrich2023fair}.
The models are trained on large amounts of data from the web~\cite{schuhmann2022laion}, granting them the inherent capacity to generate content that may contravene societal norms, including the creation of inappropriate material like pornography~\cite{schramowski2022safe}, or content that violates law such as child abuse.
More alarmingly, the inadvertent generation of inappropriate content is precipitated by spurious correlations within these models. Harmless prompts can lead to the creation of decidedly objectionable content \cite{bianchi2023easily,friedrich2023fair}. A prime example of this phenomenon lies in the correlation between specific phrases and the perpetuation of stereotypes, such as the connection between mentions of ethnicity and economic status. For example, an increase of the concept `\textit{black person}' may inadvertently amplify the appearance of the concept `\textit{poverty}.'

Conversely, methods like \model~also possess the potential to mitigate bias and inappropriateness, a prospect highlighted by prior research~\cite{friedrich2023fair, desimone2023whatis}, e.g.~through dataset augmentation~\cite{rosenberg2023unbiased}. Furthermore, established strategies offer means to mitigate the generation of inappropriate content \cite{schramowski2022safe, gandikota2023erasing} that could be deployed in tandem with \model. 
%
In summary, we advocate for a cautious approach to the utilization of these models, recognizing both the risks and promises they bring to the realm of AI-powered image editing.

\section{Further Examples on \model~Properties}\label{app:properties}
As discussed in Sec.~\ref{sec:properties}, \model~versatility benefits from re-sampling to provide variations of edits. The example in Fig.~\ref{fig:app_variations} demonstrates the additional control non-deterministic variations provide to the user, which can select the preferred interpretation of the edit instruction. 

The precision and versatility of \model~further benefit from the fact that the magnitude of an editing concept in the output scales monotonically with the edit scale $s_e$. In Fig. \ref{fig:app_monotonicity}, we can observe the effect of increasing $s_e$. Both for positive and negative guidance, the change in scale
correlates with the strength of the smile or frown. Consequently, any changes to the input image
can be steered intuitively using the edit guidance scale.

\section{Experimental Details} \label{app:expdetails}
\noindent 
Subsequently, we provide further details on the experiments presented in the main body of the paper. We first provide information on the reconstruction and runtime experiments (Sec.~\ref{sec:properties}), followed by the masking evaluation (Sec.~\ref{sec:masking}) and multi-conditioning experiments (Sec.~\ref{sec:experiments}). Details on the user study are independently described in App.~\ref{app:user_study}.
All experiments were performed using the respective diffusers\footnote{\tiny\url{https://huggingface.co/docs/diffusers}} implementation (version 0.20.2) with Stable Diffusion 1.5. 

\subsection{Properties}
First, we go into detail on the reconstruction and runtime experiments presented in Tab.~\ref{tab:properties}.
\subsubsection{Reconstruction Error}
Since the Stable Diffusion VAE already induces errors when reconstructing images, we considered the RMSE over the 64x64 latent image instead. We randomly sampled 100 images from the 2017 COCO validation dataset, which we attempted to reconstruct using the default configuration of each method as described in the respective paper or implementation. For methods that could potentially benefit from a descriptive target prompt of the input image, we considered an empty prompt, COCO caption as prompt, and unconditioned generation (no CFG) and reported the best score. Below, we outline the configuration for each method. \\ \\

\noindent
\begin{tabularx}{\linewidth}{l X}

\makecell[l]{\textbf{\model~(Ours)}~\& \\ \textbf{Edit-friendly} \\ \textbf{DDPM \cite{hubermanspiegelglas2023edit}:}}   &  Perfect reconstruction for any hyperparameter combination. The error induced by machine precision is inconsequential. \\ \\
\textbf{Imagic} \cite{kawar2023imagic}:   &  1000 embedding learning steps w/ learning rate $2e-3$ and 1500 model tuning steps w/ learning rate $5e-7$. Target prompt is the original image caption. We used 50 generation steps with $\alpha$ and guidance scale $0.0$\\ \\
\textbf{Pix2Pix-Zero} \cite{parmar2023zeroshot}: & Inversion with 100 steps, no CFG, $\lambda=20$ for auto correction and KL divergence, and $5$ regularization and auto correction steps, respectively. 100 inference steps with cross-attention guidance $0.0$ and no CFG. Source and target embeddings are null-vectors. \\ \\

\end{tabularx}
\noindent
\begin{tabularx}{\linewidth}{l X}
\textbf{DDIM Inversion:}     & 1000 inversion steps and 50 generation steps. Both without classifier-free guidance \\ \\
\textbf{SDEdit} \cite{meng2022sdedit}: & 40 diffusion steps (strength 0.8 at 50 default
steps) with no CFG. \\ \\
\end{tabularx}

\noindent
Notably, the small difference between DDIM and Pix2Pix-Zero only holds for the pure reconstruction of an image. Previous research has shown \cite{mokady2023nulltext} that for DDIM inversion, the accumulated error increases drastically when using classifier-free guidance. Since CFG is necessary for editing, the "reconstruction" portion during editing becomes worse for DDIM and remains stable for Pix2Pix-Zero.
\subsubsection{Runtime}

For runtime measurements, we consider the wallclock runtime on a dedicated NVIDIA A100-SXM4-40GB GPU. As a proxy task, we considered applying an \textit{´oilpainting'} style to a photograph. We only measured the inversion and generation loops, discarding any I/O or other processing. For each method, we considered 100 runs with hyperparameters based on the respective paper/official implementation, as outlined below. \\

\noindent
\begin{tabularx}{\linewidth}{l X}

\textbf{\model~(Ours)}: & 20 inversion and 20 generation steps with threshold $\lambda=0.1$ and skip for $t=0.75T$.\\ \\
\textbf{Edit-friendly DDPM \cite{hubermanspiegelglas2023edit}:}   & 100 inversion steps and 64 generation steps (i.e. 36 skip steps) \\ \\
\textbf{Imagic} \cite{kawar2023imagic}:   &  1000 text embedding optimization steps, 1500 model finetuning steps, 50 inference steps \\ \\
\textbf{Pix2Pix-Zero} \cite{parmar2023zeroshot}: & 100 steps at inversion and inference. 5 regularization steps and auto correction steps for each inversion step. \\ \\
\textbf{DDIM Inversion:}     & 1000 inversion steps (w/o CFG) and 50 generation steps  \\ \\
\textbf{SDEdit} \cite{meng2022sdedit}: & 40 diffusion steps (strength 0.8 at 50 default
steps). \\ \\
\end{tabularx}

\noindent
\begin{tabularx}{\linewidth}{l X}

\end{tabularx}
An interesting observation is the fact that \model, runs faster than SDEdit although both perform 40 diffusion steps overall. However, SDEdit requires 80 total U-Net evaluation (unconditioned and conditioned for each step) step, whereas \model~only requires 60 (unconditioned at each inversion step and unconditioned + conditioned at each inference step). Performing 2 evaluations of the U-Net is significantly slower than 1 evaluation even if performed as a batch. 

\subsection{Implicit Mask Quality}
We have already provided detailed information on the experiment in Sec.~\ref{sec:masking}. For further reference, we note that after removing duplicate/ambiguous objects the dataset contains 4983 images and 29307 segmentation objects.
\subsection{Multi-conditioning}
The multi-conditioning experiment presented in Sec.~\ref{sec:mult_cond}, used the following attributes: 

\begin{itemize}
  \setlength\itemsep{-0.2em}
    \item glasses
    \item smile
    \item hat
    \item wavy hair
    \item earrings 
\end{itemize}

We used the first 100 images in CelebA that were labeled to not contain any of the five target attributes. Seeds were chosen at random but kept fixed across all experiments. 
For the LPIPS and CLIP scores, we relied on the default implementation from torchmetrics\footnote{\tiny\url{https://lightning.ai/docs/torchmetrics/stable/}}. Consequently, we used the AlexNet variant with mean reduction and the original ViT-L/14 CLIP checkpoint from OpenAI\footnote{\tiny\url{https://huggingface.co/openai/clip-vit-large-patch14}}. For the CLIP scores, we calculated a dedicated score for each of the 3 applied edits and considered the mean for each image. 

The hyperparameter variations of each method were run as a grid search over the hyperparameter ranges listed below. Other parameters were kept at their default values.  For each method, we ran a grid search over a wider range of parameters to identify reasonable boundaries and subsequently discarded edge values leading to drops in performance. \\

\noindent
\begin{tabularx}{\linewidth}{l X}

\textbf{\model~(Ours)}: & Skip between 0.2 and 0.3, Guidance scale between 10.0 and 15.0, Threshold between 0.7 and 0.9\\ \\

\end{tabularx}

\noindent
\begin{tabularx}{\linewidth}{l X}
\textbf{Edit-friendly DDPM \cite{hubermanspiegelglas2023edit}:}   & Skip steps between 20 and 40, Guidance scale between 10.0 and 15.0 \\ \\
\textbf{DDIM Inversion:}     & Guidance Scale between 1.0 and 15.0  \\ \\
\textbf{Imagic} \cite{kawar2023imagic}:   &   Guidance Scale between 2 and 6, and $\alpha$ between 0.1 and 1.3  \\ \\
\textbf{Pix2Pix-Zero} \cite{parmar2023zeroshot}: &  Guidance Scale between 1.0 and 10.0 and cross guidance scale between 0 and 0.15 \\ \\
\textbf{SDEdit} \cite{meng2022sdedit}: & Guidance scale between 5.0 and 10.0 and strength between 0.2 and 0.8 \\ \\
\end{tabularx}


\section{TEdBench++}\label{app:tedbench++}

\noindent We propose TEdBench++\footnote{\tiny \url{https://figshare.com/s/7adc2b0fe1e0388dd99f}}, a revised version of TEdBench \cite{kawar2023imagic} which sets a new standard for benchmarking real text-based image editing. It is publicly available, including original images, edit instructions, and edited images with \model~for benchmarking new methods. Figs.~\ref{fig:creative_examples} and \ref{fig:failure2} as well as Tab.~\ref{tab:tedbench} demonstrated our generated images with \model~to improve upon the previous SOTA method, Imagic, setting a new reference for benchmarking. 
Next to providing better-edited images, we also addressed several inconsistencies in the target texts and missing tasks.

\begin{figure}
    \centering
    \includegraphics[width=0.75\linewidth]{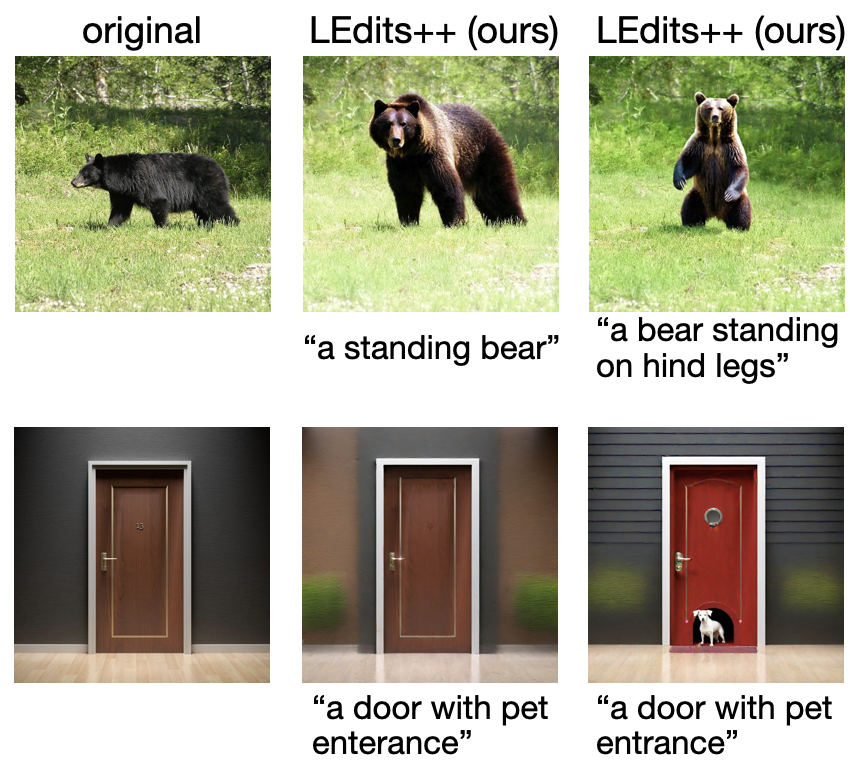}
    \caption{Exemplary inconsistencies in TEdBench and their corrections in TEdBench++. Left is the original image, and in the middle/right are images edited with \model~for the edit instructions above. All parameters (seed, etc.) are the same. As can be seen, the edit success heavily depends on the clear and correct writing of the words. In the middle column, it does not work (ambiguous (top) and misspelled (bottom)), whereas the edit is successful in the right image (clear and correctly spelled).}
    \label{fig:tedbench++corrections}
\end{figure}

We show several inconsistencies of TEdBench in Fig.~\ref{fig:tedbench++corrections}. First, we corrected ambiguous text prompts such as a \textit{standing} animal that is already standing Fig.~\ref{fig:tedbench++corrections} (top). This applied to multiple images (horse, cat, bear, etc.). Instead, we propose ``an \{animal\} standing on hind legs'' to specify the target text and thus ask for a clear but more challenging edit.
Second, we correct for misspellings such as ``enterance'' which should be ``entrance'' instead. While this may appear negligible, DMs' tokenizers may provide completely different tokens in these cases: e.g., one token for the correct word but three tokens for the misspelled word. 
In Fig.~\ref{fig:tedbench++corrections}, we show the impact of the corrections on the edit success on \model. Although we use the exact same parameters (seed, etc.), the edit of the original (left) to the middle fails, whereas it is successful for the corrected prompts (right). This way, we provide a higher-quality benchmark.

Further, we added novel tasks to the benchmark, making it more challenging and accounting for a broader range of tasks. In Fig.~\ref{fig:tedbench++}, we show examples of the new tasks we added: i) multi-editing (adding multiple concepts at the same time), ii) object removal (removing an object while staying consistent with the background and overall image composition), iii) style transfer (changing the whole image, i.e. all pixels without changing the overall image composition or object, only their style appearance), and iv) complex replacements (adding and removing multiple concepts at the same time).

With these extensions, we improve on the previous benchmark and propose a higher-quality version. This way, we hope to benefit the research community and set up a new standard for benchmarking text-based real-image editing techniques. 

\section{User Study} \label{app:user_study}
\noindent
Next to evaluating with automated metrics such as CLIP and LPIPS scores, we also conducted a study with human evaluators. We focus the user study on TEdBench(++). First, we describe the experimental details for generating the images for the study. Then, we describe the setup of the user study.

\paragraph{Experimental details}
We followed the approach of Kawar \textit{et al.}~\cite{kawar2023imagic} and generated images for several seeds and hyperparameters and hand-selected the best fitting image (exemplary grid search shown in Fig.~\ref{fig:gridsearch}). Notably, we evaluated only three seeds, whereas Kawar \textit{et al.} evaluated eight seeds. Furthermore, we limited the grid search to a decent but small range for each hyperparameter. 

For \model~(with SD1.5 and SD-XL), we set the number of diffusion steps fixed to 50 steps and grid-searched skip $[0.0,0.1,0.2,0.4]$, masking threshold $[0.6,0.75,0.9]$, and guidance scale $[10, 15]$. As a result, we evaluated 72 images (= 3 seeds $\times$ 4 skips $\times$ 3 thresholds $\times$ 2 scales) per benchmark sample. All other hyperparameters correspond to the default values of the diffusers implementation. Consequently, the generated images with \model~could be even further improved when evaluating for more hyperparameters, e.g.~more seeds (also see open question discussion on seed in App.~\ref{app:limitations}).

For Imagic with SD1.5, we relied on 3 seeds and 50 diffusion steps, too. 
We grid-searched the guidance scale $[5.0,7.5,10.0]$ and alpha value $[0.4, 0.5, 0.6, 0.7, 0.8, 0.9, 1, 1.2, 1.4, 1.6, 1.8, 2.0]$. As a result, we evaluated 108 images (= 3 seeds $\times$ 3 scales $\times$ 12 alphas) per benchmark sample. The remaining setup and parameters correspond to the default values of the original Imagic implementation with Imagen \cite{kawar2023imagic}. For Imagic with Imagen, we had to rely on their curated outputs of TEdBench since the model is not publicly available.

\begin{figure}
    \centering
    \includegraphics[width=\linewidth]{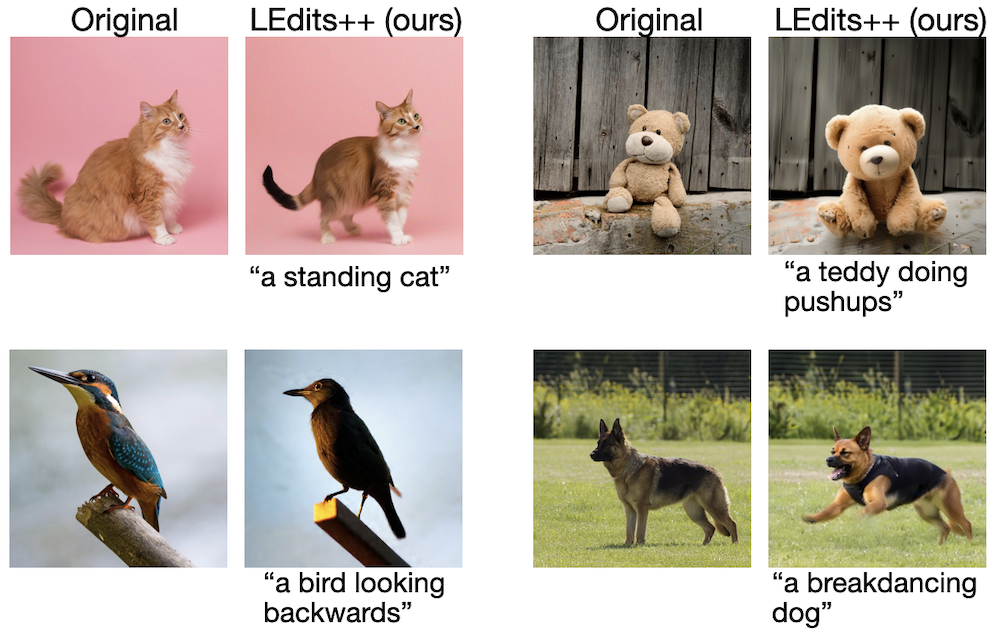}
    \caption{Failure cases of \model~on TEdBench}
    \label{fig:failure}
\end{figure}

\begin{figure*}
    \centering
    \begin{subfigure}[t]{0.47\textwidth}
        \centering
        \includegraphics[width=0.8\linewidth]{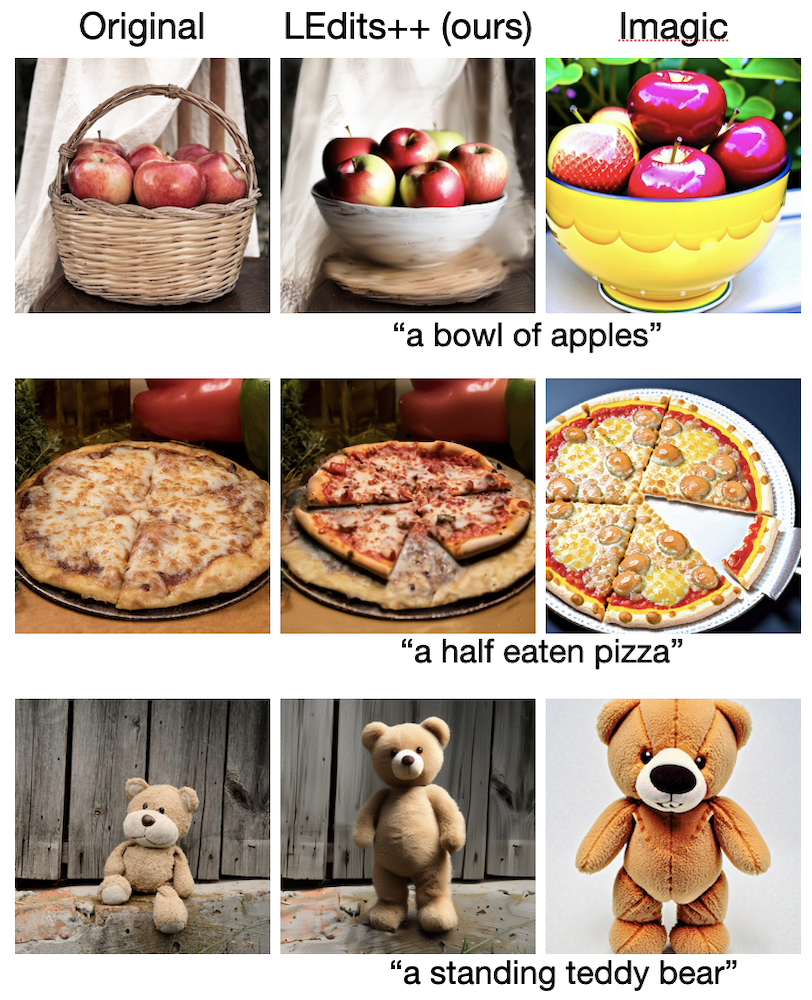}
        \caption{Image editing or generating a new image?}
        \label{fig:failure2a}
    \end{subfigure}
    \begin{subfigure}[t]{0.47\textwidth}
        \centering
        \includegraphics[width=0.8\linewidth]{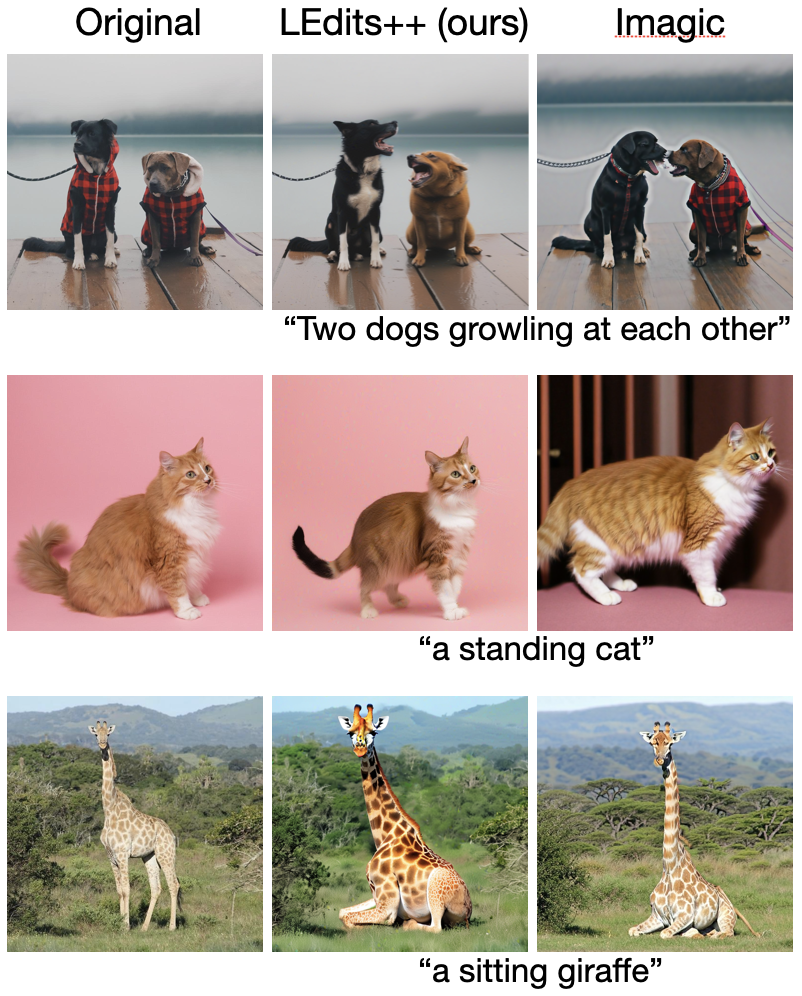}
        \caption{Coherence Trade-off: compositional robustness vs.~object identity}
        \label{fig:failure2b}
    \end{subfigure}
    \caption{Comparing failure cases of \model~and Imagic on TEdBench.}
    \label{fig:failure2}
\end{figure*}
\begin{figure*}
    \centering
    \begin{subfigure}[b]{\textwidth}
        \centering
        \includegraphics[width=0.95
        \textwidth]{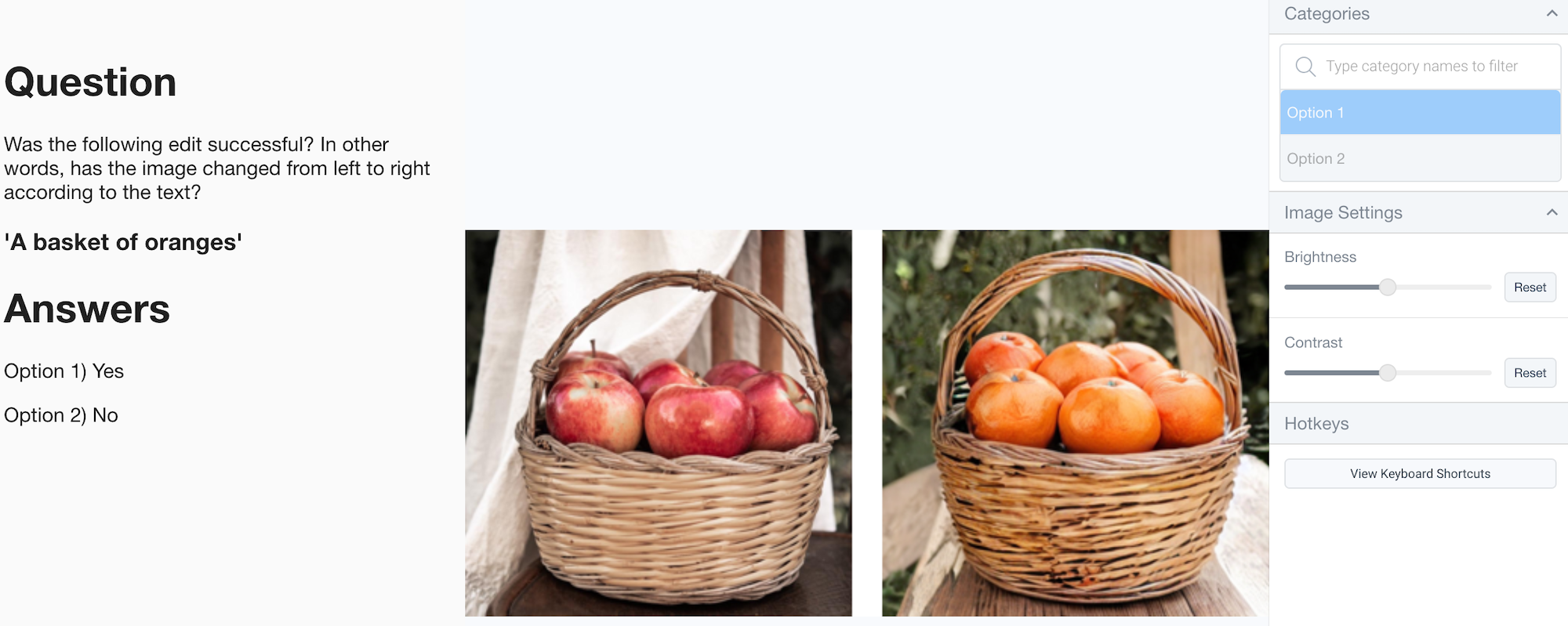}
        \caption{Setup for user study: ``was the editing successful?''}
        \label{fig:userstudysetup}
    \end{subfigure}
    \begin{subfigure}[b]{\textwidth}
        \centering
        \includegraphics[width=0.95
        \textwidth]{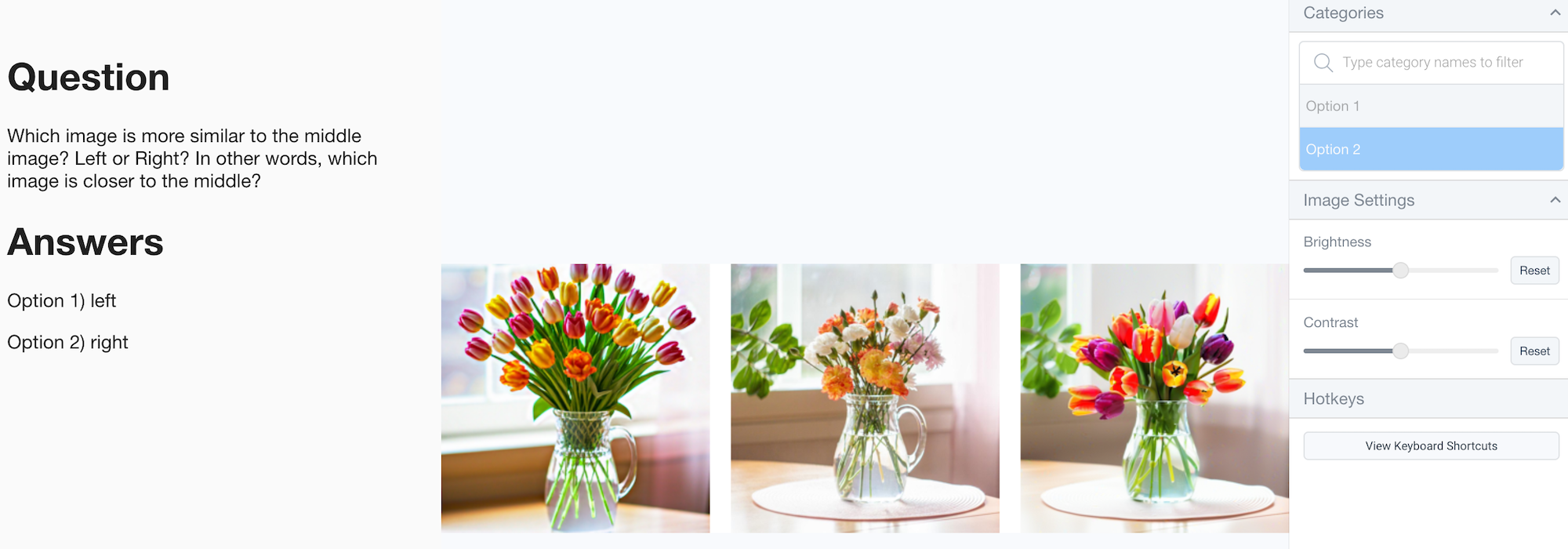}
        \caption{Setup for user preference study: ``which edited image is closer to the original image?''}
        \label{fig:userstudypreferencesetup}
    \end{subfigure}
    \caption{User study setups for both user studies conducted. The first user study evaluates the edit success of an image editing method. The second user study evaluates the user preference between two image editing methods regarding image-to-image similarity.}
    \label{fig:userstudies}
\end{figure*}
\begin{figure*}
    \centering
    \includegraphics[width=\textwidth]{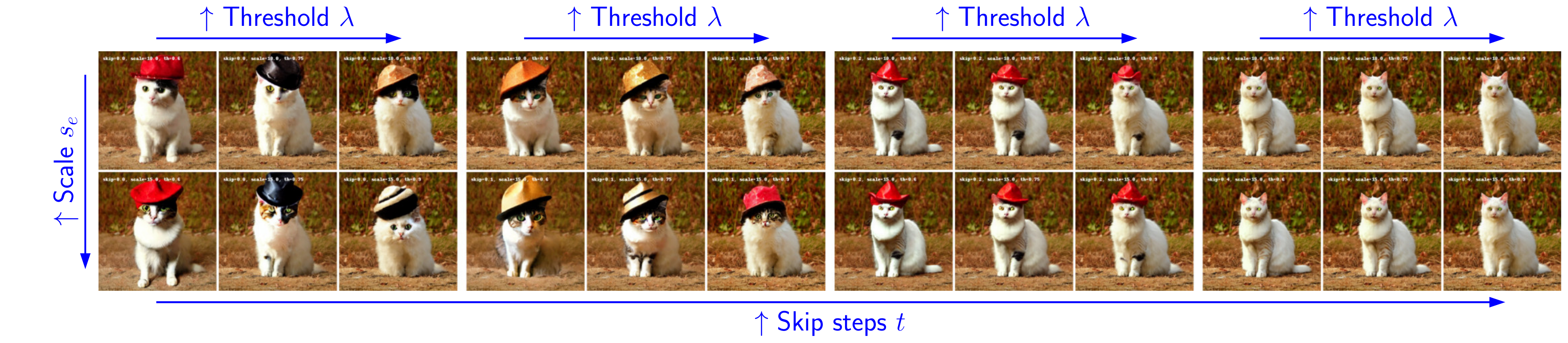}
    \caption{Grid search results for \model~for TEdBench(++). We show the grid search for the image ``cat.jpeg'' from the benchmark and the target text ``a cat wearing a hat''. From left to right the parameters are increased. We searched three parameters, the skip steps, the guidance scale, and the masking threshold. As can be seen, the stronger the parameters, the more changes are made to the image. On the other hand, too weak parameters do not change the image, i.e.~do not realize the target text. This highlights the trade-off between edit success and preservation of the image composition and object identity.
    All images are generated for the same seed (for TEdBench(++) we evaluated 3 seeds per benchmark entry, whereas Imagic used 8 seeds). }
    \label{fig:gridsearch}
\end{figure*}

\paragraph{User study setup}
For the actual user studies, we chose the following setups (cf.~Fig.~\ref{fig:userstudies}) on the platform \href{https://thehive.ai}{thehive.ai}.
Users had to pass a qualifying test, and during the actual labeling, 15\% of the tasks a user saw were honeypot tasks (sanity check). Only if the qualifier test was passed error-free and the honeypot accuracy was permanently above 95\%, we accepted the given answers to ensure high-quality evaluations. Users could zoom in/out and change several image parameters, such as brightness and contrast to further enable a high-quality assessment. 

The first study (see Fig.~\ref{fig:userstudysetup}), which we also describe in the main text in Tab.~\ref{tab:tedbench}, evaluates the success rate of an editing technique on TEdBench(++). To this end, we asked users to assess the overall success of edits, i.e., if an edit instruction was faithfully realized for a given input image. Fig.~\ref{fig:userstudysetup} shows the setup, in which a user had to choose between two options, whether the edit instruction has been realized successfully or not. The setup consisted of a general question-and-answer setting for all examples alongside a specific edit text for each image pair. The original image (always left) and the edited image (always right) were shown in the center. Outputs from \model~and Imagic were interleaved at random. The result is given in Tab.~\ref{tab:tedbench} in which \model~clearly outperforms Imagic for both underlying DMs and both versions of the benchmark benchmarks.

We also conducted a second user study. In this study, we asked the user to assess the image similarity of two methods to a reference image, see Fig.~\ref{fig:userstudypreferencesetup}. 
With this human preference study, we investigate the image-to-image similarity after editing, i.e., if the edited images still look similar to the original one. 
Participants were shown the input image (middle) and were asked to choose the better editing result from one of the two methods (left and right), using the common practice of Two-Alternative Forced Choice (2AFC). The methods were randomly switched between left and right to avoid confounding factors.
To this end, we compared \model~(with SD-XL) and Imagic (with Imagen) on TEdBench. For this comparison, we considered only images where both methods were labeled successful in the prior study. 
In that study, users preferred \model~over Imagic with 60\% preference. As outlined previously, this again emphasizes the precision of our method, which preserves the overall image composition and results in high-quality edits. Yet, the preference seems smaller than the results with the LPIPS scores. We found this to be an artifact of imprecise user study design. A preference setup generally suffers from bias such as subjects replacing general questions with more specific ones \cite{schwarz1999how} (e.g.~``which is more similar?'' might be replaced by ``in which did the main object stay the same, regardless of the background?'' or ``in which are the background and overall image composition better preserved regardless of the main object?''). Hence, it is difficult to draw exact conclusions from this study, but a clear trend is still visible. Moreover, as shown in the main text in Tab.~\ref{tab:tedbench}, we computed LPIPS scores, which further clarify the results of the user study. Additionally, we broadly discussed the similarity trade-off between object identity/coherence and overall image composition in the limitation sections of the main body (Sec.~\ref{sec:discussion}) and appendix (App.~\ref{app:limitations}).

\section{Limitations and Further Discussion}
\label{app:limitations}
\noindent 
In the following, we extend the discussion of the main body with further examples and questions.

\paragraph{Model Dependency.} In general, we observed the editing success to be dependent on the underlying DM. In Fig.~\ref{fig:comp_model_giraffe}, we show that the generation of a \textit{sitting giraffe} depends on the underlying DM. For both editing techniques, the weaker SD1.5 variant fails but the more advanced variant succeeds.
Upon further investigation, we realized that SD1.5 is incapable at all of generating images of \textit{sitting} giraffes. In Fig.~\ref{fig:sd15_giraffe_grid}, we show exemplary images for the text prompt ``a sitting giraffe'' (we generated 100 and all showed the same result) and can see that none is actually sitting. In contrast, SD-XL is able to output images of \textit{sitting} giraffes (cf.~Fig.~\ref{fig:sdxl_giraffe_grid}) and consequently enables \model~to perform the desired edit. This emphasizes the importance of choosing an apt underlying DM for real image editing and motivates future research to develop more powerful DMs from which editing techniques will benefit, too.

\paragraph{Failure Cases and Open Questions.}
In the following, we want to touch upon open questions and failure cases. In Fig.~\ref{fig:failure}, we show several failure cases of \model. In the first case, the cat is indeed edited from a sitting to a standing cat. Yet, the identity of the cat has changed, i.e., the shape of the tail and the fur color have changed. We show further examples in Fig.~\ref{fig:failure2}. However, defining what makes an edit \textit{acceptable} remains challenging and may differ between users, applications, and context.
In general, however,  there are two reasons for the discussed limitations. First, \model~limits its edits to the identified relevant regions. Consequently, the background and image composition will be preserved, but the edit within this region depends on various factors, including hyperparameter strength, underlying DM, and random seed.
Second, a lack of descriptiveness of the editing prompt with respect to the specific identity of an object can lead to changes thereof.
Especially if generic terms such as `cat' are used to edit the image. To guarantee the preservation of the object's identity, methods like Textual Inversion \cite{gal2022textual}, or Break-a-Scene \cite{avrahami2023break} could be employed. 

The other three examples in Fig.~\ref{fig:failure} show failure cases of \model beyond changes of object identity, i.e., cases in which the edit instruction is not or not sufficiently realized. There are several reasons for such failures, including a general lack of concept understanding in the underlying DM (discussed above), incorrect masking of the relevant region, wrong hyperparameter choices, or challenging prompts. For example, what exactly is a ``breakdancing dog'' supposed to look like? We even considered removing this entry from the benchmark but found it very challenging at the same time and, therefore, kept it. 
Moreover, we found the edit success rate and quality to depend on the used seed. This is in line with current work on the impact of the used seed on the diffusion process \cite{samuel2023generating}. Samuel~\textit{et~al.} propose a new method to identify fitting seeds, which could be applied to \model~as well to find satisfying editing seeds.

\begin{figure*}[h!]
    \centering
    \includegraphics[width=0.65\linewidth]{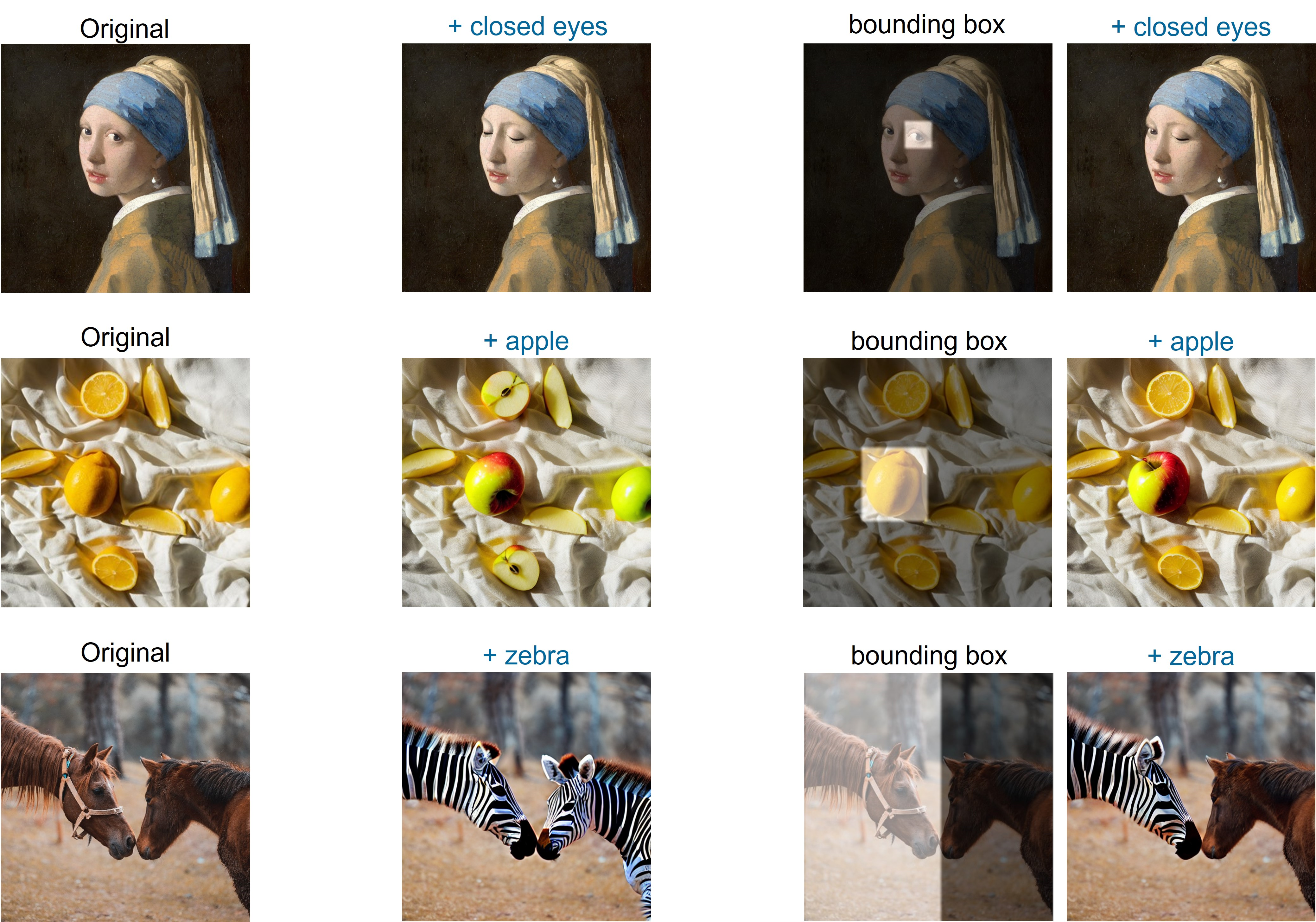}
    \caption{Customized and complex image editing with \model~for individual user masking. \model~can be easily extended with user masks to facilitate user preferences. The first column shows the original images and the second and fourth show the edited images with \model~and the target text above. The third column shows user-provided masks, marking the relevant region for the edit instruction. In the second column, \model~uses implicit masking as implemented in our default approach, and in the fourth column \model~uses the explicit user-provided masks from the third column.}
    \label{fig:user_mask}
\end{figure*}

\paragraph{Masking and User Interaction}
The automatically-inferred implicit masks allow for easy use of \model~without users tediously providing masks. Nonetheless, user intentions are diverse and cannot always be automatically inferred.
Sometimes, individual user masks provide better control over the editing process. Such user masks can be easily integrated into \model. In Fig.~\ref{fig:user_mask}, we show customized image editing with individual user masking. \model~can be easily extended with user masks to facilitate user preferences. Sometimes \model's implicit masks do not meet user preferences or it is difficult to textually describe the relevant image region. Next to using dedicated models to obtain image masks, users can simply provide their own masks. In our setup, we did not evaluate this scenario as it drastically increases resources in terms of computation or human labor. Yet, it can be easily integrated into \model. Fig.~\ref{fig:user_mask} shows the original image can be edited well with \model. Moreover, dedicated user masks help focus on a specific image region. This is specifically helpful for logical and compositional instructions (current models struggle with ``left''/``right'' etc.), for one specific object if multiple are present (one specific ``orange'' from several ones), and for both combined. This way, \model~stays lightweight in its default with implicit masking, but can still and easily handle user masks and thereby implement individual user experience.

\begin{figure*}[h!]
    \centering
    \includegraphics[width=0.9\linewidth]{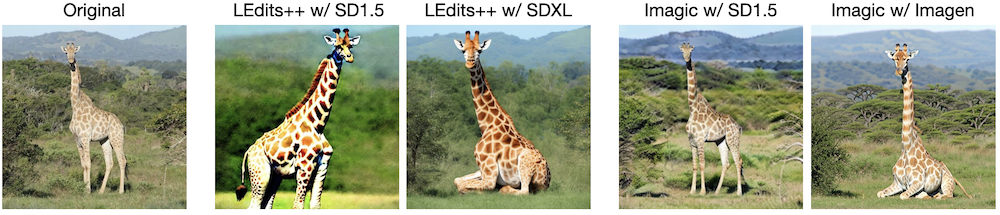}
    \caption{Impact of underlying DM. The original image (leftmost) should be edited with the target text ``a photo of a sitting giraffe''. The edit success depends on the underlying DM: with SD1.5 it fails whereas it works fine with more advanced DMs (SD-XL and Imagen). This holds for both methods \model~and Imagic.}
    \label{fig:comp_model_giraffe}
\end{figure*}

\begin{figure*}[h!]
    \centering
    \includegraphics[width=0.9\linewidth]{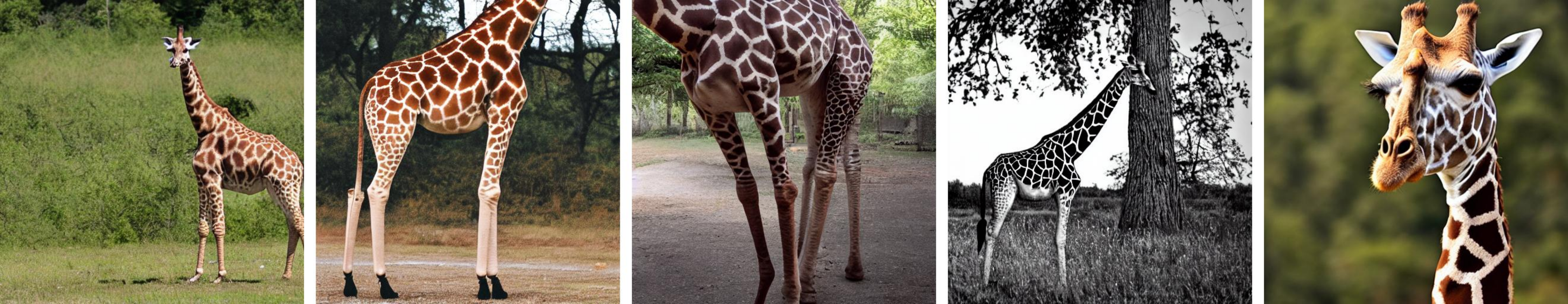}
    \caption{Generated Images with SD1.5 for ``a photo of a sitting giraffe''. The model consistently fails to generate a giraffe in that specific pose. }    \label{fig:sd15_giraffe_grid}
\end{figure*}

\begin{figure*}[h!]
    \centering
    \includegraphics[width=0.9\linewidth]{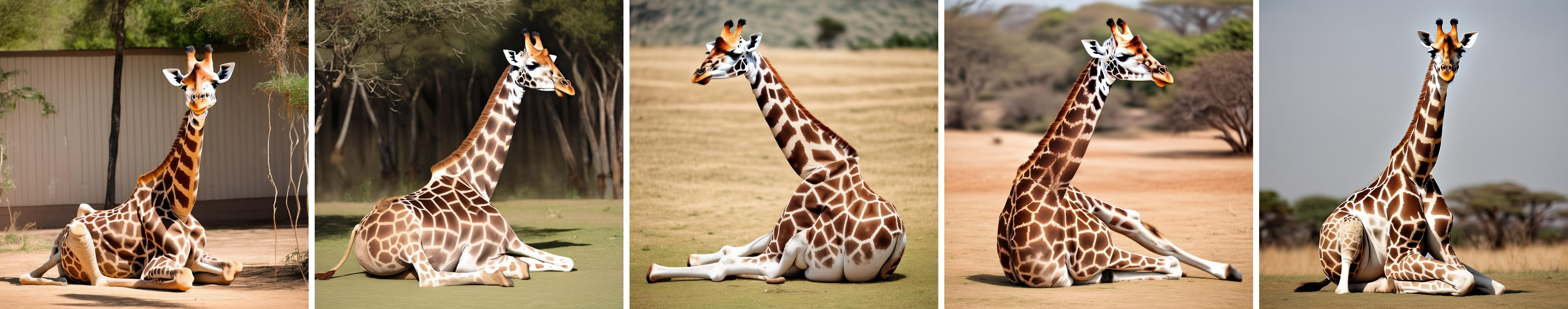}
    \caption{Generated Images with SDXL for ``a photo of a sitting giraffe''.}
    \label{fig:sdxl_giraffe_grid}
\end{figure*}

\section{Further Results}
\label{app:further_results}
\noindent
Subsequently, we present further results, qualitative examples, and visual ablations. 

\subsection{Qualitative examples}
We show further results in Fig.~\ref{fig:animalvariation}. We remove ``cat'' and add a diverse set of animals instead. Interestingly, this works for a variety of animals, that share no or only little similarity, such as ``flamingo'' or ``parrot''. Furthermore, the newly occurring background is inpainted semantically sensible, too. Additionally, we show more qualitative examples in Fig.~\ref{fig:further_examples}.

\subsection{Ablations}
In Fig.~\ref{fig:gridsearch}, we show an ablation of \model~for TEdBench(++). This grid search illustrates the impact of different hyperparameters on the trade-off between edit success and the preservation of the overall image composition/ object identity.

\subsection{Semantic Grounding Ablations}
\begin{figure*}
    \centering
    \includegraphics[width=0.7\linewidth]{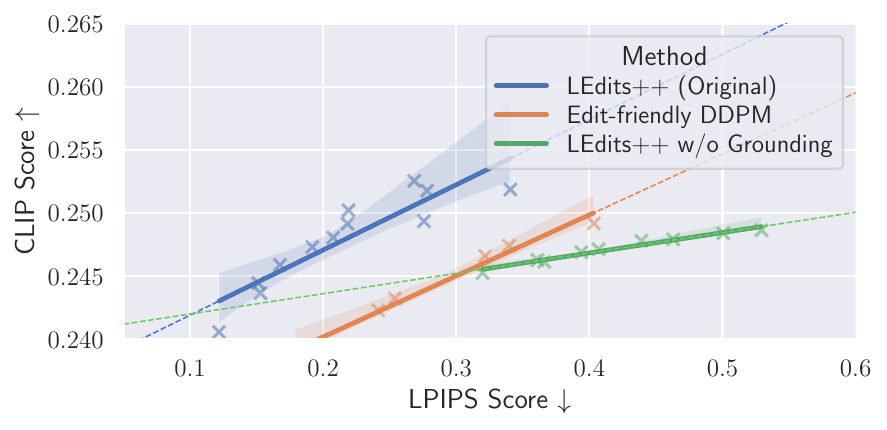}
    \caption{Semantic Grounding Ablations. Semantic grounding is an essential component of \model~that helps preserving overall image composition and realizing concise edit instructions. (Best viewed in color)}
    \label{fig:semanticgrounding}
\end{figure*}

We performed extensive ablations on semantic grounding by re-running \model~on Sec.~\ref{sec:masking}'s benchmark without any grounding. The results in Fig.~\ref{fig:semanticgrounding} show that \model~will still achieve strong instruction alignment (high CLIP score) without grounding, but semantic masking is key to keeping the generated image similar to the input (low LPIPS score). Moreover, grounding allows for a clearer trade-off between instruction alignment and image similarity in the first place. We believe these ablations foster a deeper understanding of the importance of semantic grounding in the \model~pipeline.

\subsection{Explanatory Visualization of Monotonicity}
\begin{figure*}
    \centering
    \includegraphics[width=.7\linewidth]{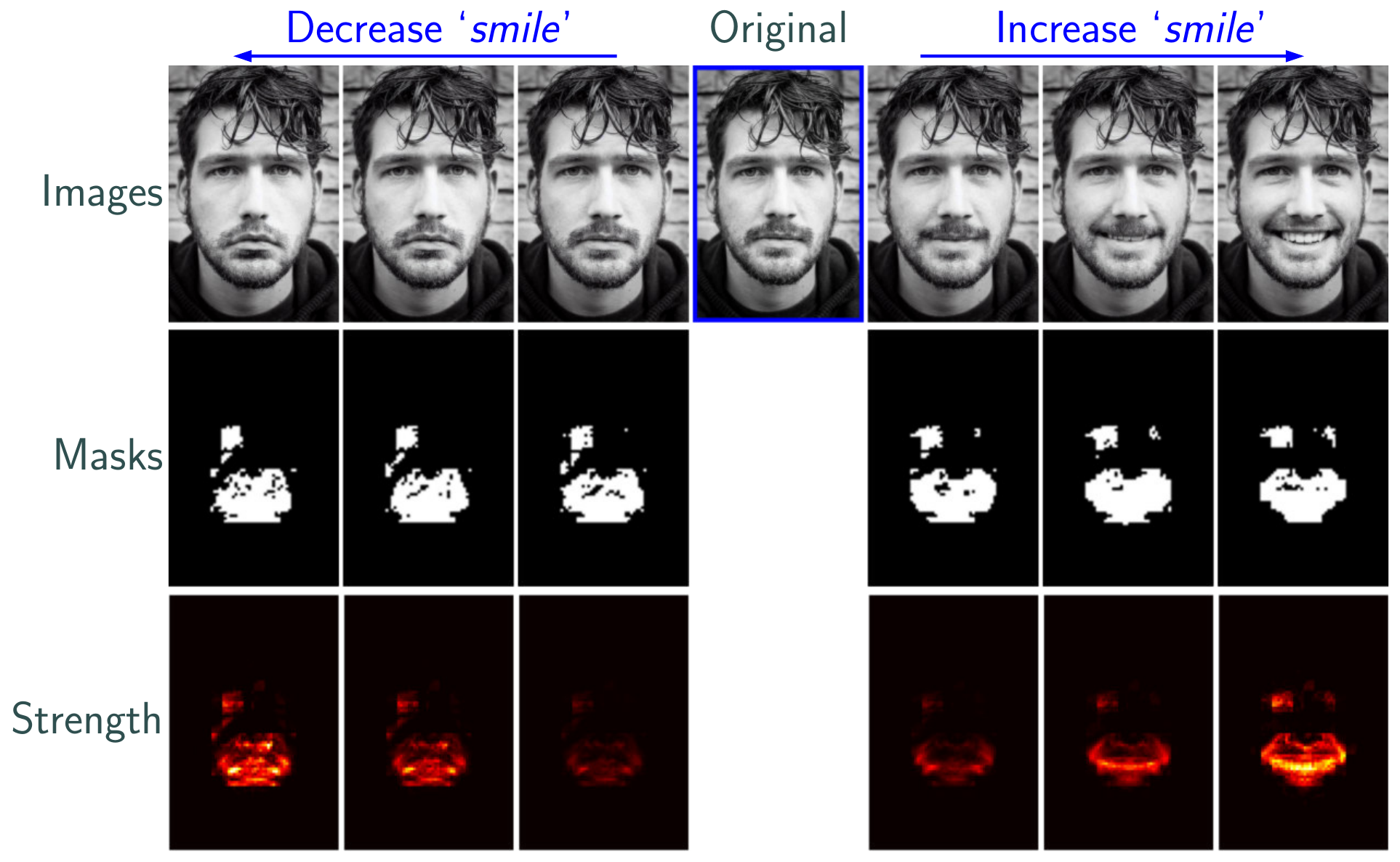}
    \caption{Masking and guidance scale. Within the identified edit regions, the magnitude of applied changes correlates directly with the changing scale. (Best viewed in color)}
    \label{fig:maskstrength}
\end{figure*}

The monotonicity of the \model~guidance scale is an important contribution of the method.
Importantly, the inferred masks are mostly isolated from changes to the guidance scale. In the example shown in Fig.~\ref{fig:maskstrength}, we would expect the masks for `\textit{smiling}' to always target the area around the mouth and eyes. Within these identified regions, the magnitude of applied changes correlates directly with the changing scale, as evident from the provided heatmap. Here, we can also observe that different areas are prioritized depending on the magnitude of the change. The initial focus is clearly on the mouth, with strong changes in the eyes only appearing for larger scales.

\subsection{Masking and Artifacts}
\begin{figure*}
    \centering
    \includegraphics[width=0.7\linewidth]{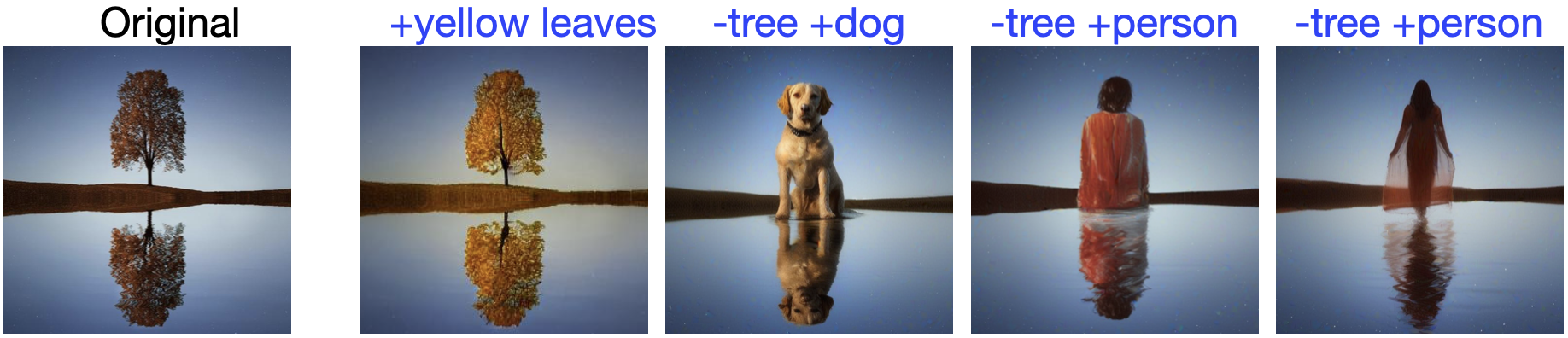}
    \caption{\model~can easily handle complex edits such as reflections. (Best viewed in color)}
    \label{fig:reflection}
\end{figure*}

\begin{figure*}
    \centering
    \includegraphics[width=0.7\textwidth]{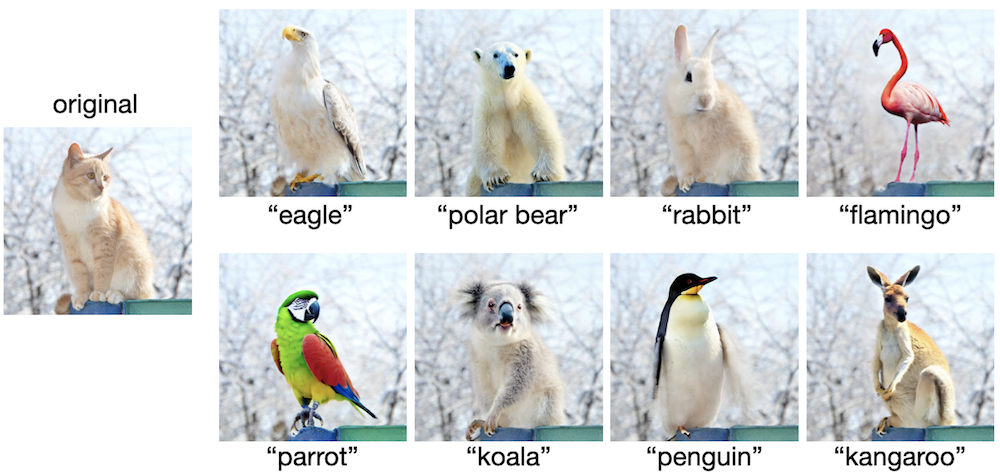}
    \caption{Object replacement with \model. The leftmost image is the original image and the other images are edited with \model~and the target text below. We apply diverse replacements of the main object with the overall image composition being preserved. Interestingly, the background is filled and interpolated very well, e.g.~for ``flamingo'' or ``parrot''. (Best viewed in color)}
    \label{fig:animalvariation}
\end{figure*}

\begin{figure*}
    \centering
    \includegraphics[width=0.9\textwidth]{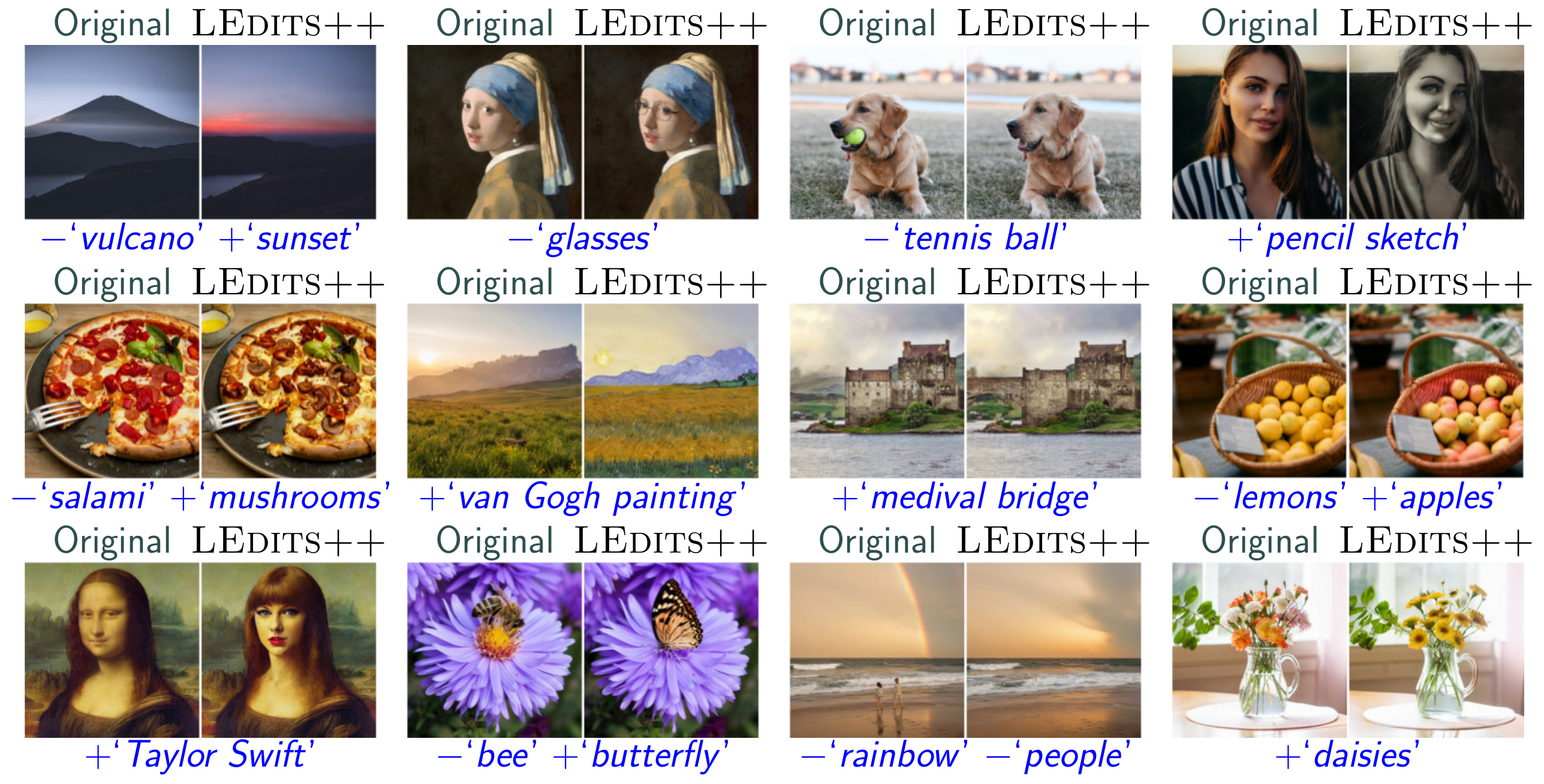}
    \caption{Further qualitative examples of image editing with \model, highlighting its versatility and precision  (Best viewed in color)}
    \label{fig:further_examples}
\end{figure*}

\model~can faithfully edit reflections/shadows of objects, even with masking (cf.~Fig.~\ref{fig:basic_examples} \& \ref{fig:reflection}). This ability strongly depends on the underlying diffusion model. In the example in Fig.~6b, the underlying diffusion model simply failed to correctly correlate the couple and their shadow/reflection. However, in Fig.~\ref{fig:reflection} where SDXL is applied, the reflection is edited as well.

\end{document}